\documentclass[10pt,twocolumn,letterpaper]{article}

\usepackage[pagenumbers]{cvpr}

\usepackage{graphicx}
\usepackage{amsmath}
\usepackage{amssymb}
\usepackage{booktabs}
\usepackage{times}
\usepackage{epsfig}
\usepackage{listings}
\usepackage{multirow}
\usepackage[table,xcdraw]{xcolor}
\usepackage{verbatim}
\usepackage{color, colortbl}
\usepackage{bm}
\usepackage{enumitem}
\usepackage{pifont}
\usepackage[accsupp]{axessibility}
\usepackage{array}
\usepackage{float}
\usepackage[linesnumbered,ruled,vlined]{algorithm2e}
\usepackage[pagebackref,breaklinks,colorlinks]{hyperref}
\usepackage[font=small,labelfont=small]{caption}

\definecolor{Gray}{gray}{0.9}
\newcommand{\cmark}{\ding{51}}
\newcommand{\xmark}{\ding{55}}
\newcommand\minisection[1]{\vspace{1mm}\noindent \textbf{#1}}
\newcommand{\model}{BeMapNet\xspace}
\newcommand{\modelbf}{\textit{\textbf{BeMapNet}}\xspace}
\newcommand{\bev}{\textit{BEV}\xspace}
\newcommand{\Bezier}{B\'ezier\xspace}
\newcommand{\sota}{\textit{SOTAs}\xspace}
\newcommand{\LD}{$lane$-$divider$\xspace}
\newcommand{\PC}{$ped$-$crossing$\xspace}
\newcommand{\RB}{$road$-$boundary$\xspace}
\newcommand{\hdmap}{\textit{HD-map}\xspace}
\newcommand{\twod}{$2$\textit{D}\xspace}
\newcommand{\threed}{$3$\textit{D}\xspace}
\newcommand{\nuscene}{\textit{NuScenes}\xspace}
\newcommand{\red}[1]{\textcolor{red}{#1}}

\newcommand{\bluecell}[1]{\cellcolor[HTML]{DAE8FC}{#1}}
\newcommand{\greycell}[1]{\cellcolor[HTML]{F0FFF0}{#1}}

\def\confName{CVPR}
\def\confYear{2023}

\begin{document}

\title{End-to-End Vectorized HD-map Construction with Piecewise \Bezier Curve}
\author{
	Limeng Qiao \quad Wenjie Ding \quad Xi Qiu \textsuperscript{\ding{41}} \quad Chi Zhang \\
	\textit{MEGVII Inc.} \\
	{\tt\small \{qiaolimeng, dingwenjie, qiuxi, zhangchi\}@megvii.com}
}
\maketitle

\begin{abstract}
	Vectorized high-definition map (\hdmap) construction, which focuses on the perception of centimeter-level environmental information, has attracted significant research interest in the autonomous driving community. 
	Most existing approaches first obtain rasterized map with the segmentation-based pipeline and then conduct heavy post-processing for downstream-friendly vectorization.
	In this paper, by delving into parameterization-based methods, we pioneer a concise and elegant scheme that adopts unified piecewise \Bezier curve. 
	In order to vectorize changeful map elements end-to-end, we elaborate a simple yet effective architecture, named Piecewise B\'ezier \hdmap Network (\modelbf), which is formulated as a direct set prediction paradigm and postprocessing-free.
	Concretely, we first introduce a novel IPM-PE Align module to inject \threed geometry prior into \bev features through common position encoding in Transformer.
	Then a well-designed Piecewise \Bezier Head is proposed to output the details of each map element, including the coordinate of control points and the segment number of curves.
	In addition, based on the progressively restoration of \Bezier curve, we also present an efficient Point-Curve-Region Loss for supervising more robust and precise \hdmap modeling.
	Extensive comparisons show that our method is remarkably superior to other existing \sota by $18.0$ \textit{mAP} at least \footnote{\footnotesize \url{https://github.com/er-muyue/BeMapNet}}.
\end{abstract}

\footnotetext{\text{\ding{41}} ~Corresponding Author}

\section{Introduction}
\label{sec:intro}
As one of the most fundamental components in the auto-driving system, high-definition map contains centimeter details of traffic elements, vectorized topology and navigation information, which instruct ego-vehicle to accurately locate itself on the road and understand what is coming up ahead.
At present, conventional \textit{SLAM-based} solutions \cite{zhang2014loam, shan2018lego, shan2020lio} have been widely adopted in practice.
Yet, due to dilemmas of high annotation costs and untimely updates, the offline approach is gradually being replaced by the learning-based online \hdmap construction with onboard sensors.

\begin{figure}[t]
	\vspace{-0.cm}
	\begin{center}
		\includegraphics[width=0.96\linewidth]{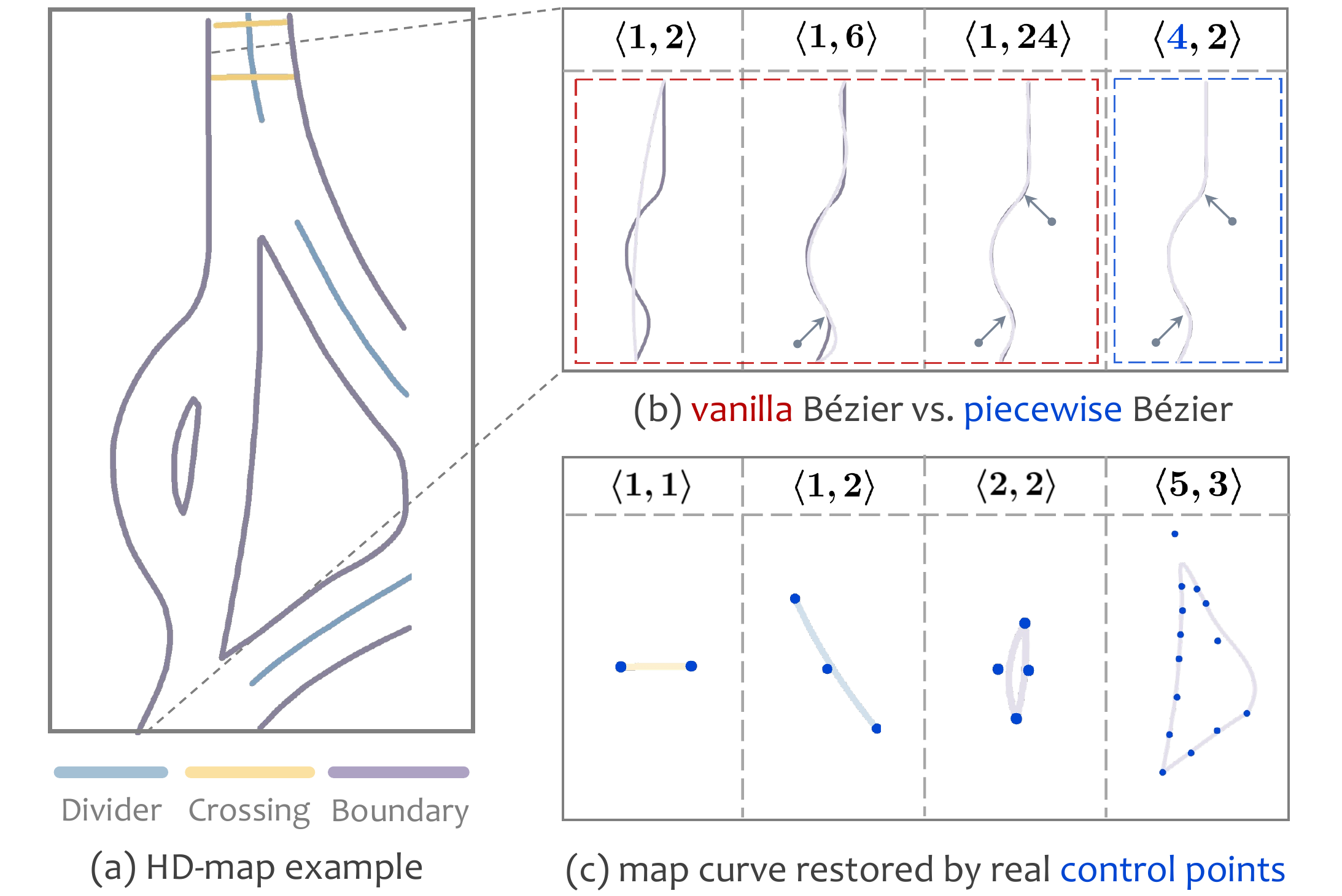}
	\end{center}
	\vspace{-0.52cm}
	\caption{
		Illustration of our motivation for piecewise  \Bezier curve, termed as \bm{$\langle k, n \rangle$}, where \bm{$k$} is the piece number and \bm{$n$} is the degree. 
		\textbf{\textit{Fig.(a)}} is a real \hdmap case from \nuscene. 
		\textbf{\textit{Fig.(b)}} compares the difference between vanilla and piecewise \Bezier curve through the same map element, where the light purple is the restored curve with \Bezier process.
		The last is more efficient than previous ones with reducing the number of control points by \bm{$64\%$} in this case.
		\textbf{\textit{Fig.(c)}} illustrates that piecewise \Bezier curve can model arbitrary-shaped curves. Note the blue circles denote actual control points.
	}
	\vspace{-0.6cm}
	\label{fig:motivation}
\end{figure} 


The deep-based paradigm of online \hdmap building is gradually developing, but it still faces two main challenges: 
\\ 
\noindent 
\textbf{\textit{1)}} \textit{modeling instance-level vectorized \hdmap end-to-end.}
Most existing works construct \hdmap by rasterizing \bev (bird-eye-view) maps into semantic pixels with segmentation \cite{li2022hdmapnet, peng2022bevsegformer}, which not only lacks the modeling of instance-level details, but also requires heavy post-processing to obtain vectorized information. As a sub-task, lane detection makes a relatively better advance for this issue, that is, in addition to segmentation-based methods \cite{neven2018towards, pan2018spatial, zheng2021resa}, there are also point-based \cite{li2019line, Tabelini_2021_CVPR} and curve-based \cite{liu2021end,feng2022rethinking} schemes. 
However, compared to the simple lane scenario, \hdmap contains more shape-changeful elements, so such methods cannot be directly adopted into the \hdmap construction.
\\
\noindent
\textbf{\textit{2)}} \textit{performing \twod-\threed perspective transformation efficiently.}
Obtaining \threed-\bev perception from multi-view $2$\textit{D} images is an essential step for building \hdmap, which is mainly divided into three ideas, \ie geometric priors \cite{reiher2020sim2real}, learnable parameters \cite{pan2020cross, peng2022bevsegformer}, and a combination of the two \cite{philion2020lift, huang2021bevdet}.
Note the assumptions of geometry-based methods often do not conform to the actual situation, leading to such schemes are less adaptable, while learning-based methods require a large amount of labeled data to generalize across various scenarios.
Combining the above two branches not only has multi-scenario scalability, but also reduces the demand for annotated data, has attracted increasing research interest.

To the best of our knowledge, the curve parameterization construction of \hdmap in the \bev space is vacancy and no one has explored it.
Based on the widely used \Bezier curve, which is mathematically defined by a set of control points, we pioneer to devise a concise and elegant \hdmap scheme that adopts \textbf{\textit{piecewise}} \Bezier curve, where each map curve is divided into multiple \bm{$k$} segments and each segment is then represented by a vanilla \Bezier curve with degree \bm{$n$},  hence denoted as \bm{$\langle k, n \rangle$}. 
Despite \bm{$\langle 1, n \rangle$} is enough to express any map element with infinite \bm{$n$} in theory, more complex curve tends to require higher degree, meaning that there are more control points need to be modeled, which is shown in Fig.\ref{fig:motivation}.
The proposed piecewise strategy allows us to parameterize a curve more compactly with fewer control points and higher capacity, which is extremely scalable and robust in practice.

\begin{figure*}[htb]
	\begin{center}
		\includegraphics[width=1.0\linewidth]{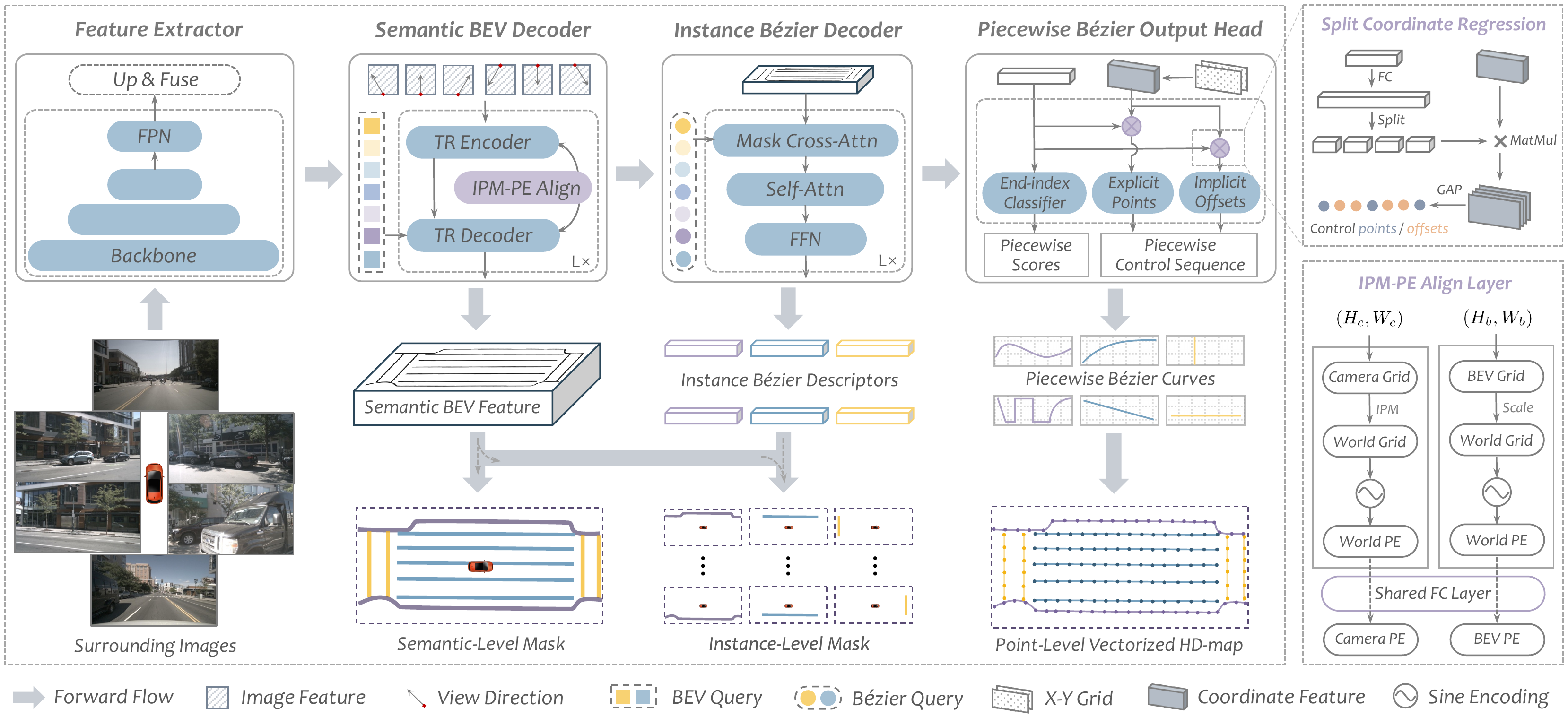}
	\end{center}
	\vspace{-0.57cm}
	\caption{
		The architecture of our proposed \modelbf, containing four primary components  for extracting progressively richer information, \ie \textit{image-level} multi-scale features, \textit{semantic-level} \bev feature, \textit{instance-level} curve descriptors, and \textit{point-level} \Bezier control sequence. 
		Right-top: The blue and orange circles represent explicit control points and implicit control offsets respectively. Note that the term \textit{GAP} is \textit{global average pooling} and \textit{MatMul} is the \textit{matrix multiplication}.
		Right-bottom: $(H_c, W_c)$/$(H_b, W_b)$ is the shape of image/\bev feature.
	}
	\vspace{-0.44cm}
	\label{fig:framework}
\end{figure*}

Inspired by the above motivations, we propose an end-to-end vectorized \hdmap construction architecture, named as \textit{Piecewise} \textbf{B\'e}zier \textit{HD-\textbf{map}} \textbf{Net}work (\modelbf).
The overall framework is  illustrated in detail in Fig.\ref{fig:framework}, which streamlines the architecture into four primary modules for gradually-enriched information, \ie feature extractor shared among multi-view images, semantic \bev decoder for \twod-\threed perspective elevation, instance \Bezier decoder for curve-level descriptors,  and piecewise \Bezier head for point-level parameterization. 
To be concrete, we first introduce a novel \textit{IPM-PE Align module}  into Transformer-based decoders, which injects \textit{IPM} (inverse perspective mapping) geometric priors into \bev features via \textit{PE} (position encoding) and hardly adds any parameters except a \textit{FC} layer.
Secondly, we further design a \textit{Piecewise \Bezier Head} for dynamic curve modeling with adopting two branches as classification and regression, where the former classifies the number of piece to determine the curve length and the latter regresses the coordinate of control points to determine the curve shape.
Lastly, we present an \textit{Point-Curve-Region Loss} for robust curve modeling by supervising restoration information as a progressive manner.
Since it is modeled as a sparse set prediction task and optimized with a bipartite matching loss, our method is postprocessing-free and high-performance.

\noindent The main contributions of our approach are three-folds:
\begin{itemize}[leftmargin=12pt,topsep=2pt, parsep=-1pt]
	
	\item We pioneer the \modelbf for concise and elegant modeling of \hdmap with unified piecewise \Bezier curve.
	
	\item We elaborate the overall end-to-end architecture with innovatively introducing \textit{IPM-PE Align Module}, \textit{Piecewise \Bezier Output Head} and well-designed \textit{PCR-Loss}.
		
	\item \modelbf is remarkably superior to \sota on existing benchmarks, revealing the effectiveness of our approach.
	
\end{itemize}

\section{Related Work}
\label{sec:related}

\subsection{Vectorized \textbf{\textit{HD-map}} Construction}

\hdmap is precise at centimeter-level and consists of vectorized details not normally present on standard map \cite{liu2020high}, which highly aggravates the difficulty of obtaining accurate annotations in practice. 
Most previous methods focus on SLAM \cite{dube2017online, lee2013robust, mendes2016icp, shan2020lio, yang2018robust, jiao2018machine} with exploring LiDAR points of the environment, involving large-scale data acquisition and labor-intensive annotations. 
Recently, more and more researchers formulate the process of building \hdmap as a segmentation task \cite{chen2017deeplab, he2017mask, kirillov2019panoptic, long2015fully} from various sensors, such as cameras \cite{pan2020cross, zhou2022cross, Hu_2021_ICCV, li2022bevformer, philion2020lift, yang2021projecting, mattyus2015enhancing, mattyus2016hd} and  LiDAR \cite{yang2018hdnet, besl1992method}.
HDMapNet \cite{li2022hdmapnet} extracts features from the observations of multi-modalities and groups semantic rasterized maps with heuristic post-processing for vectorized results.
Without any further vectorization, \cite{peng2022bevsegformer} only obtains semantic map representations from a Transformer-based camera-to-BEV module.
Different from the above segmentation-based frameworks, our proposed \modelbf adopts the parameterization-based paradigm and constructs instance-level vectorized \hdmap end-to-end with a multi-view and camera-only manner, which is a more flexible and scalable solution for downstream tasks.

\subsection{Structure Modeling of Geometric Data}
\hdmap contains various kinds of map elements, including \LD, \PC, \RB, \etc, which are typically regarded as geometric data, \eg points, polygons and curves.
To the best of knowledge, there are two mainstream exploration directions for deep geometric modeling, one is point-based, including uniform-points and keypoints estimation, and the other is curve-based, including polynomial curve and B\'ezier curve. 
Taking the most related lane detection as an example,  LineCNN \cite{li2019line} presents a novel line proposal unit to detect lanes as a set of points, and later  LaneATT \cite{Tabelini_2021_CVPR} represents a lane by equally-spaced \twod-points that achieves good performance. 
GANet \cite{wang2022keypoint} focuses on keypoint estimation and association with adaptively lane feature aggregator.
HDMapGen \cite{mi2021hdmapgen} proposes a hierarchical point generative model for producing high-quality HD lane map. 
As for cruve-based, \cite{van2019end} and \cite{tabelini2021polylanenet} directly predict the polynomial coefficients with a differentiable least-squares fitting module and simple \textit{FC} respectively.
Recently, based on the DETR \cite{carion2020end}, LSTR \cite{liu2021end} presents a query-based detector to decode poly-parameters of a lane end-to-end.
Apart from this, some studies also adopt B\'ezier curve for geometry, such as text boundary \cite{liu2020abcnet},  center-line segment \cite{can2021structured} and lane curve \cite{feng2022rethinking}. 
Unlike these approaches of addressing simple-shape elements, our method innovatively employs piecewise B\'ezier curves without any complicated geometric assumptions to parameterize arbitrary-shape geometry in \hdmap scenarios.

\subsection{Multi-view Camera-to-BEV Transformation}
In most \threed research studies, obtaining high-quality \bev representations from multi-view camera features is the top priority of various domains, such as \threed object detection \cite{philion2020lift, wang2021fcos3d, wang2022detr3d, liu2022petr, huang2021bevdet}, motion prediction\cite{djuric2020uncertainty, hu2021fiery, jin2017predicting, lee2017desire, walker2014patch} and map construction \cite{li2022hdmapnet, peng2022bevsegformer, li2022bevformer, pan2020cross}.
IPM \cite{reiher2020sim2real} is the most basic and straightforward method with a homography transformation, which is precisely calculated by camera parameters. 
Without adopting the \threed geometry prior, some methods \cite{pan2020cross, can2021structured, peng2022bevsegformer} directly leverage learnable parameters to complete the perspective transformation.
VPN \cite{pan2020cross} and Neat \cite{chitta2021neat} both utilize a \textit{FC} layer to transform the image features into the \bev space, and  \cite{peng2022bevsegformer} further designs a multi-camera deformable attention unit based on transformer.
However, in order to obtain more explicit \bev representations, more and more researches argue that the geometry priors has great advantages for model convergence and performance.
Based on VPN, \cite{li2022hdmapnet} fuse the multi-camera \bev spaces with the camera poses.
LSS \cite{philion2020lift} and BEVDet \cite{huang2021bevdet} build the connection between camera-view and \bev based on the depth distribution estimation.
DETR3D \cite{wang2022detr3d} manipulates predictions directly in \threed space with linking \threed positions to \twod spaces and \cite{chen2022persformer} generates \bev features by regarding camera parameters as a reference.  
PETR \cite{liu2022petr} encodes the \threed information as position embedding and conducts query decoding on position-aware features.
In addition to introducing the \threed geometry prior, we further performs position embedding alignment for transforming the camera features and \bev features into the world coordinate system.

\section{Method}
\label{sec:method}

\subsection{Problem Formulation}
\label{sec:formulation}
\begin{figure}[t]
	\vspace{-0.cm}
	\begin{center}
	\includegraphics[width=0.98\linewidth]{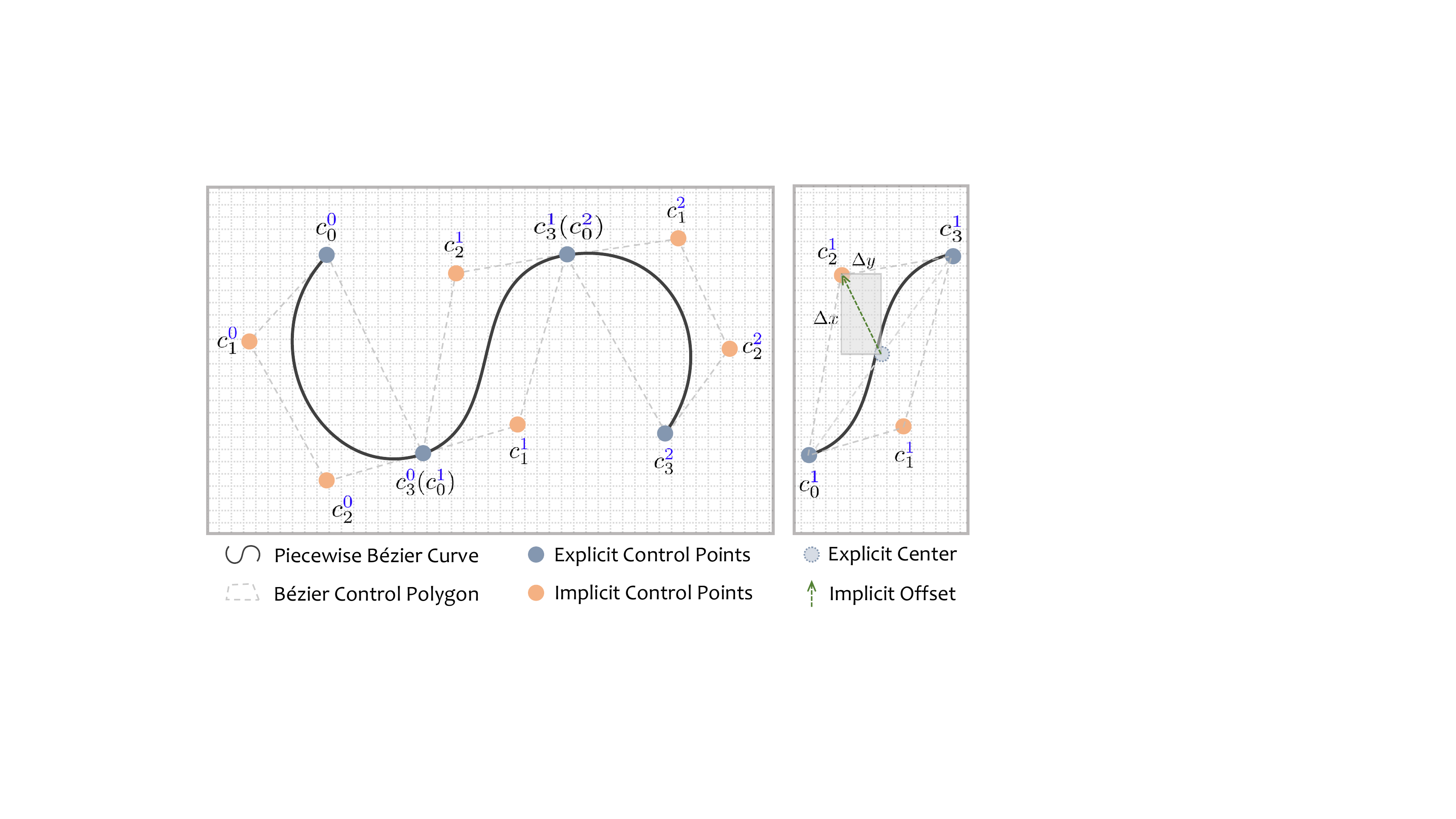}
	\end{center}
	\vspace{-0.56cm}
	\caption{Illustration of a \textit{3-pieces 3-order} piecewise b\'ezier curve, denoted as \bm{$\langle 3, 3 \rangle$} for the left part. For each segment, we model the implicit control point coordinates by an offset $(\Delta{x}, \Delta{y})$ from the center of explicit control points as shown on the right part.}
	\vspace{-0.59cm}
	\label{fig:pbc}
\end{figure}

\vspace{-0.1cm}
\minisection{Preliminary on Piecewise B\'ezier Curve.}
\label{para: pbcurve}
A B\'ezier curve is a parametric curve which is formulated by a set of ordered control points $c_0$ through $c_n$ as, 
\begin{align}
	\label{eq:bezier}
	p(t) = \sum_{i=0}^{n} b_{i, n}(t)c_i, \ t \in [0, 1]
\end{align}
where $n$ is the degree of the curve and $b_{i, n}(t)$ is known as Bernstein basis polynomial of degree $n$,
\begin{align}
	\label{eq:bernstein}
	b_{i, n}(t) = {n \choose i} t^i(1-t)^{n-i}, \ i = 0, \dots, n
\end{align}
According to Eq.\ref{eq:bezier}, we know the first and last control points are endpoints of the curve, \ie $p(0) = c_0, \ p(1) = c_n$. 
Then we define a piecewise B\'ezier curve \bm{$\langle k, n \rangle$} consists of $k$ segments, each of which is an $n$\textit{-order} B\'ezier and two consecutive segments satisfy the positional continuity condition as $p^{j}(1) = p^{j+1}(0)$, where $j$ is the segment \textit{id} varying from $0$ to $k$-$2$. 
Hereafter, given a piecewise B\'ezier curve as shown in Fig.\ref{fig:pbc}, we denote its ordered control points as,
\begin{align}
	\label{eq:bezier-defination}
    \mathbb{C}=\{ c^j_i \in \mathbb{R}^2 \vert i \in [0, n], j \in [0, k-1], c^j_n = c^{j+1}_0\}  %
\end{align}
Moreover, due to the first and last control points of each segment are always on the curve, we naturally term these points that retain a distinct significance as \textit{explicit control points} $\mathbb{C}^E$ and the remaining as \textit{implicit control points} $\mathbb{C}^\mathcal{I}$, where $\mathbb{C} = \mathbb{C}^E \cup \mathbb{C}^\mathcal{I}$, $\vert \mathbb{C}^E \vert = k+1$ and $\vert \mathbb{C}^\mathcal{I} \vert = nk-k$. 

\minisection{Vectorized \hdmap Modeling.}
By sequentially sampling points along the elements in \hdmap, each object can be structured as an open-shape curve as a vectorized manner.
To uniformly parameterize these various shape-changeful map elements, including points, polygons and curves, we employ the piecewise B\'ezier curve proposed in Eq.\ref{eq:bezier-defination} and formalize a vectorized \hdmap as a piecewise B\'ezier curve set $\mathcal{M} = \{ \mathbb{C}_i \vert i=1, \dots, \vert \mathcal{M} \vert \}$, where $\vert \mathcal{M} \vert $ is the number of map instances. 
Our objective is to learn a model that extracts compact information from the camera sensors and predicts, for each map element, its corresponding B\'ezier control sequence, which uniquely determine the shape and position of the map curve.
In addition, for the purpose of reducing the modeling complexity of implicit control points $\mathbb{C}^\mathcal{I}$ for each segment, we  subtly generate implicit coordinates through the center of explicit control points and their relative offsets as shown in the right of Fig.\ref{fig:pbc}.  

\minisection{Piecewise B\'ezier Curve Modeling Principle.}
\label{sec: principle}
Considering the large-variety and changeful-shape of map elements, we present a \textbf{\textit{consistent-degree but dynamic-piece}} principle for modeling piecewise curve more precisely and efficiently as, 

\noindent
(1) \textbf{\textit{consistent-degree on semantic-level}}. 
Due to a B\'ezier curve of degree $k$ can be easily converted into degree $k+1$ with the same shape, \ie \textit{degree elevation }in mathematics,  the same curve may correspond to multiple B\'ezier solutions with only different degree. 
This kind of obfuscation is devastating for the network modeling.
And we argue that keeping the \textit{consistent-degree} within the same semantic category is the most efficient way to solve this dilemma.

\noindent
(2) \textbf{\textit{dynamic-piece on instance-level}}. 
Considering that the complexity of different map curve is always inconsistent, the modeling method with a fixed and large number of segments will not only cause excessive redundancy in the expression of each curve, but also lead to confusion in the definition of sub-segment, which places a certain burden on the model optimization.  
The \textit{dynamic-piece} rule demands the model to choose the number of segments dynamically and parameterize a B\'ezier curve as compactly as possible.

\subsection{Piecewise \Bezier HD-map Network}
\label{sec:bemapnet}

\subsubsection{Overall Architecture}
\label{sec:overall-arch}

The overall model architecture is illustrated in detail in Fig. \ref{fig:framework} , which streamlines the framework into four parts, namely  feature extractor, semantic \bev decoder, instance B\'ezier decoder and piecewise B\'ezier output head respectively.

\minisection{Feature Extractor. \label{minisection: Feature Extractor} }
Given surrounding images from multi-camera as inputs, a shared CNN backbone is first employed to obtain each image feature, and then these multi-scale features from different stages are fed into FPN\cite{tan2020efficientdet} to integrate rich environmental information. 
Last we upsample pyramid features to the same size and stack them together as outputs.

\minisection{Semantic \bev Decoder.}
We leverage a standard encoder-decoder paradigm based on Transformer to elevate camera-view features into canonical \bev spaces. By treating the perspective transformation as a direct set prediction task, the \bev decoder takes camera features with shape $H_c \times W_c$ and $H_q \times W_q$ learnable \bev queries as inputs, and produces \bev features $F_b \in \mathbb{R}^{C \times H_b \times W_b}$ by modeling all pairwise interactions among elements with self- and cross-attention.
Different from the region-agnostic query in DETR, we correspond each query with a $\gamma^h \times \gamma^w$ region on \bev features one-to-one, \ie $H_b \times W_b = \gamma^h  H_q \times \gamma^w W_q$.
The $F_b$ is then fed into a $1 \times 1 \ conv$ and \textit{upsample} block to obtain the final semantic mask $\mathbb{M}_{s} \in  \mathbb{R}^{U \times H_s \times W_s}$, where $U$ is the number of classes and $H_s \times W_s$ is the map shape.
In addition, we also propose a novel \textit{\textbf{IPM-PE Align Layer}} to refactor the features in different coordinate systems through the most common position encoding layer in Transformer.

\minisection{Instance \Bezier Decoder.}
So as to perform more accurate parametric modeling of each curve, we further equip an instance \Bezier decoder based on masked cross-attention \cite{cheng2022masked}.
To be concrete, given semantic features $F_b$ and learnable instance \Bezier queries $Q \in \mathbb{R} ^ {V \times C}$, where $V$ is the max number of instance, this module aggregates information and decodes \Bezier descriptors $\vec{z} \in \mathbb{R} ^ {V \times C}$, which contain key info of geometric and positional relationships between different map elements.
With extra performing matrix multiplication between  $\vec{z}$ and $F_b$, we obtain instance map mask $\mathbb{M}_z \in  \mathbb{R}^{V \times H_s \times W_s}$, which is used as the foreground mask for next decoder layer and also to perform segmentation supervision for more spatial context information.

\minisection{Piecewise B\'ezier Output Head.}
Following the principle elaborated in Sec.\ref{sec: principle}, we closely design a piecewise \Bezier output head with two core modules, \ie \textbf{\textit{Split Coordinate Regression}} and \textbf{\textit{Dynamic End-index Classification}}, which are utilized to output the coordinates of \Bezier control point sequence and the number of segments respectively. With easily recovering piecewise \Bezier curves and further introducing the bipartite matching setting, the proposed output head constructs vectorized local \hdmap very efficiently.

\subsubsection{IPM-PE Align Layer}
\label{sec:ipmpe}
As a basic module of traditional Transformer architecture, positional encoding utilizes the sequence order by injecting information about the position of tokens \cite{vaswani2017attention}, where \textit{sin-cos} and \textit{learned-based} functions are the most common practices.
Yet, for the purpose of perspective transformation between \twod camera-views and \threed \bev, we argue it is not enough to only encode the \textit{position correspondence} within single view, but it is also necessary to maintain the \textit{position consistency} relationships between two perspectives.
Thence we put forward a novel \textit{IPM-PE Align Layer} to encode the \twod-\threed geometry priors into features from different coordinate systems.
To be concrete, given a point $p^{f_i}_{c}=(u, v, 1)^{\top}$ on the $i$-th camera-view feature and its corresponding world point $p^{w}_c=(x, y, z)^{\top}$, the following mathematical equation is satisfied on the assumption of pinhole camera model,
\begin{align}
	\label{eq:ipm-pe-align-src}
	d \cdot A^{-1} \cdot p^{f_i}_{c} = K^i \cdot T^i \cdot p^{w}_c
\end{align}
where $d$ is the depth, $A$ is the transformation matrix between image-grid and feature-grid, $K^i$ and $T^i$ are $i$-th intrinsic and extrinsic matrices.
Based on the assumption of that the ground surface is flat and at a fixed height in \textit{IPM}, the world position $(x, y)$  and depth $d$ can be easily calculated through Eq.\ref{eq:ipm-pe-align-src}. Note $d < 0$ denotes the position is not valid.
As for the other branch of \bev perspective, given a point $p^{f}_{b}$ from the \bev feature $F_b$ and its world point $p^w_b$, there is usually only a scale relationship 
$\boldsymbol{\kappa} = (\kappa_{x}, \kappa_{y})$ between them,
\begin{align}
	\label{eq:ipm-pe-align-tgt}
	p^{f}_{b} = \boldsymbol{\kappa} \cdot p^{w}_b
\end{align}
Then we leverage the standard \textit{sin-cos function} to convert all these world coordinates $P^w_c$ and $P^w_b$ as position embedding $f^{pe}_c$ and $f^{pe}_b$ respectively.
Since the assumption of flat-ground and known-height usually does not hold in practice, $f^{pe}_c$ and $f^{pe}_b$ are not exactly aligned in the world coordinate system, we further adopt a shared \textit{FC} layer on $f^{pe}_c$ and $f^{pe}_b$ to perform embedding alignment and then obtain more unified position encoding, which is shown in Fig.\ref{fig:framework} in detail.

\subsubsection{Piecewise \Bezier Output Head}
\minisection{Split Coordinate Regression Head.}
Following the illustration in the upper right corner of the Fig.\ref{fig:framework}, we elaborate the forward flow of proposed regression head as four steps, 

\noindent \textit{\textbf{1)}} with $i$-th incoming \Bezier descriptor $\vec{z}_i \in \mathbb{R}^{C}$, we first convert its channel from $C$ to $u \cdot v$ with adopting a standard \textit{FC} layer and then split it into $u$ parts to get a collection of coordinate descriptors $x_i^j \in \mathbb{R}^{v}$, where $j \in [1, u]$, $u$ is the number of coordinate and $v$ is the channel of descriptor.

\noindent \textit{\textbf{2)}} given the desired output shape $(H_s, W_s)$, a hard-coded candidate coordinates grid $G \in \mathbb{R}^{2 \times H_s \times W_s}$ is generated with X-Y two channels and next a $1 \times 1 \ conv$ layer is utilized to convert $G$ to its coordinate feature $F_G \in \mathbb{R}^{v \times H_s \times W_s}$.

\noindent  \textit{\textbf{3)}} through conducting the matrix multiplication between the coordinate descriptor $x_i^j$ and feature $F_G$, the coordinate activation map $h_i^j \in \mathbb{R}^{H_s \times W_s}$ is obtained dynamically. 

\noindent  \textit{\textbf{4)}} after using a global average pooling on the spatial of $h_i^j$, the activation map is regressed to the final coordinate value. 

\minisection{Dynamic End-index Classification Head.}
Based on the prior of that the end point of a piecewise \Bezier curve must be explicit, the proposed module model the dynamic length of segments prediction task as a $N$-classification problem, where $N$ is the maximum pieces number for a certain map class.
Specifically, with using a common \textit{FC} and \textit{softmax} block, each \Bezier descriptor is naturally transformed into a $N$-dimensional probability vector, where each position denotes the score of the current index as a termination point.
This novel indeterminate length modeling greatly increases the adaptability and scalability of our proposed framework.

\subsection{End-to-End Training} 
\label{subsec:end-toend-training}

\noindent According to the matrix form of Eq.\ref{eq:bezier}, that is, $\bm{P=BC}$, where $B \in \mathbb{R}^{m \times n}$ is the Bernstein matrix, $C \in \mathbb{R}^{n \times 2}$ is control points, $P \in \mathbb{R}^{m \times 2}$ is vectorized points on the curve, $m$ is the number of points and $n$ is the degree. 
Obviously, given $m, n$, we can easily implement following procedures:

\noindent \textbf{\textit{1) vectorization}}. Given the $C$, its corresponding vectorized curve can be restored efficiently with matrix multiplication.

\noindent \textbf{\textit{2) construction}}. Given the orderly sampling points $P$, we readily construct $C$ with solving the least-squares problem, \ie $C=B^+P$ where $^+$ is the pseudo-inverse of a matrix.

\vspace{0.05cm}
\minisection{Piecewise \Bezier Ground Truth.}
The common \hdmap annotation protocol always represents a curve with a set of vectorized points. We propose to first select some annotated keypoints and then divide the curve into $k$ segments, where each is compactly modeled by an $n$-order \Bezier curve.
Following the procedure of \textbf{\textit{construction}}, we present the ground truth generation algorithm in the Algorithm \ref{algo_pbc-generation}.

\IncMargin{1em}
\begin{algorithm} [hbp]
	\SetKwData{Left}{left}\SetKwData{This}{this}\SetKwData{Up}{up} \SetKwFunction{GetBernsteinCoefficient}{GetBernsteinCoefficient}\SetKwFunction{MatrixPseudoInverse}{MatrixPseudoInverse}\SetKwFunction{CurveInterpolate}{CurveInterpolate} \SetKwFunction{ChamferDistance}{ChamferDistance} 
	\SetKwInOut{Input}{\textit{input}}\SetKwInOut{Output}{\textit{output}}
	
	\Input{Annotated points $P$, Parameters $n$, $m$, $\epsilon$}
	\Output{Piecewise B\'ezier curves $\mathbb{C}$}
	\BlankLine 
	$B$ \ \ = \GetBernsteinCoefficient($n$, $m$)\;
	$B^+$ = \MatrixPseudoInverse($B$)\;
	$l \leftarrow \vert P \vert, s \leftarrow 0$, $e \leftarrow l-1$\;
	\While{$s < e$}{
		$P^\dagger = $ \CurveInterpolate($P[s, e]$)\;
		$\mathbb{C}^\dagger = B^+ \times P^\dagger$\;
		$P^\ddagger = B \times \mathbb{C}^\dagger $\;  
		$D' = $ \ChamferDistance($P^\dagger$, $P^\ddagger$)\;
		\If{$D' < \epsilon$}{
			$\mathbb{C} \ \leftarrow \ [ \mathbb{C}, \ \mathbb{C}^\dagger]$, $s \leftarrow e$, $e \leftarrow l-1$\;
		}
		\lElse{
			$e \leftarrow e-1$
		} 
	}
	\caption{Piecewise B\'ezier Curve \textit{\textbf{GenGT}}.}
	\label{algo_pbc-generation} 
\end{algorithm}
\DecMargin{1em} 

\begin{figure}[t]
	\vspace{-0.cm}
	\begin{center}
		\includegraphics[width=0.96\linewidth]{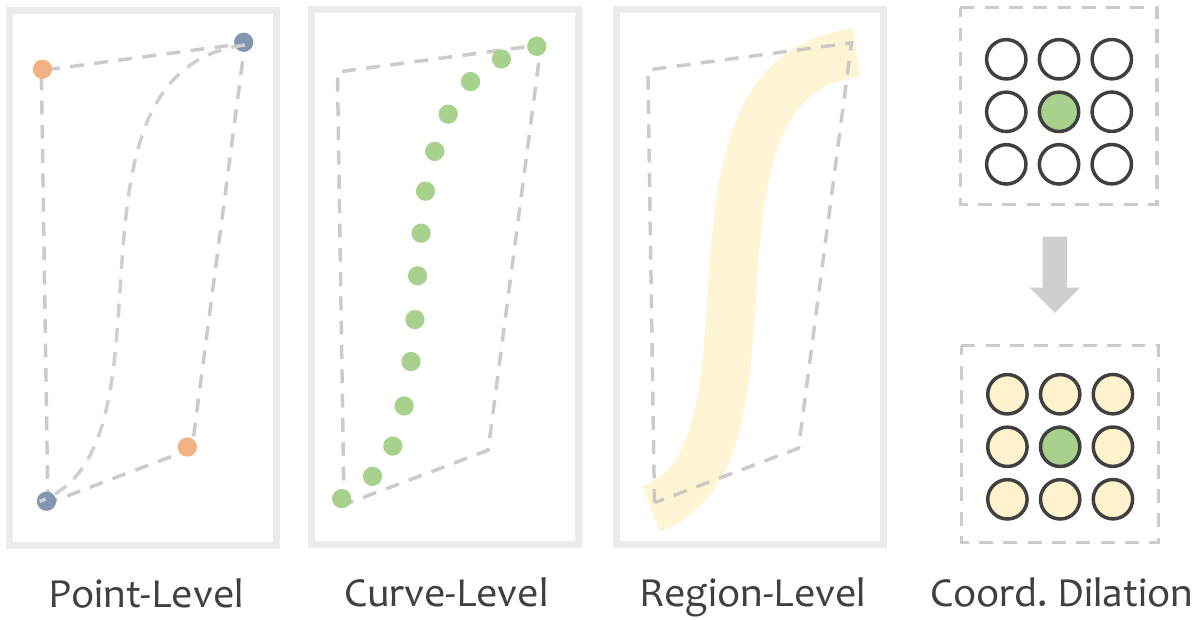}
	\end{center}
	\vspace{-0.52cm}
	\caption{
		The first three diagrams are illustrations of \textit{Point-Curve-Region Recovery Loss} for B\'ezier curve modeling.
		On the far right is a simple schematic diagram of coordinate dilation with $\omega=1$.
	}
	\vspace{-0.48cm}
	\label{fig:pcr}
\end{figure} 

\minisection{Point-Curve-Region Recovery Loss.}
Given the \Bezier control points, a novel \textbf{\textit{PCR-Loss}} is proposed to supervise the learning procedure from three progressive levels, \ie point, curve and region, which is shown in Fig.\ref{fig:pcr} in detail.

\noindent 
\textit{\textbf{1)}} \textit{point-level supervision}. 
Assume the prediction of control sequence from the output head as $\widehat{\mathbb{C}}$ and its corresponding ground truth is $\mathbb{C}$, we firstly leverage the $L1$-loss between these coordinates sequentially as, 
\vspace{-0.23cm}
\begin{align}
	\label{eq:point-loss}
	\mathcal{L}_{point} = 	\frac{1}{\vert \widehat{\mathbb{C}} \vert} \mathop{\sum} _{i=1}^{\vert \widehat{\mathbb{C}} \vert}\Vert \mathbb{C}_i - \widehat{\mathbb{C}}_i \Vert_1,
\end{align}
\vspace{-0.23cm}
\\
\textit{\textbf{2)}}  \textit{curve-level supervision.}
Due to the parametric property of \Bezier curve that a slight deviation of control points $\widehat{\mathbb{C}}$ might cause a great change of the curve, we further restore the control sequence $\widehat{\mathbb{C}}$ (or $\mathbb{C}$) to curve sequence $\widehat{\mathcal{P}}$ (or $\mathcal{P}$) with the \textbf{\textit{vectorization}} process and adopt the $L1$-loss as,
\vspace{-0.23cm}
\begin{align}
	\label{eq:curve-loss}
	\mathcal{L}_{curve} = 	\frac{1}{\vert \widehat{\mathcal{P}} \vert} \mathop{\sum}_{i=1}^{\vert \widehat{\mathcal{P}} \vert} \Vert \mathcal{P}_i - \widehat{\mathcal{P}}_i \Vert_1
\end{align}
\vspace{-0.23cm}
\\
\textit{\textbf{3)}}  \textit{region-level supervision.}
For digging out more intuitive supervision information, we further recover curve $\widehat{\mathcal{P}}$ from the form of discrete coordinates into region mask.
First of all, we introduce a coordinate dilation operator, which takes the current coordinate $p_{ij}$ as the center and then generates $(2\omega+1)^2$ surrounding coordinates $p_{\alpha\beta}$ with the giving dilation width $\omega$, where $\alpha \in [i-\omega, i+\omega]$ and $\beta \in [j-\omega, j+\omega]$.
Secondly, we conduct the coordinate dilation on each point from curve $\widehat{\mathcal{P}}$ and then obtain dilated curve points sequence $\widehat{\mathcal{P}}^{\sharp}$, which can be regarded as the foreground coordinates of the predicted mask $\widehat{M}$, \ie $\mathcal{S}(\widehat{M}, \widehat{\mathcal{P}}^{\sharp}) = \boldsymbol{1}$, where $\mathcal{S}$ is the grid sampling operation to compute the output values using $\widehat{M}$ and point coordinates $\widehat{\mathcal{P}}^{\sharp}$ from grid.
Finally, after preforming the grid sampling $\mathcal{S}$ on ground truth map mask $M$ with the same dilated prediction curve, we then bridge a segmentation-base region supervision between  $M$ and $\widehat{\mathcal{P}}^{\sharp}$, which can be formulated mathematically as, 
\vspace{-0.14cm}
\begin{align}
	\label{eq:region-loss}
	\mathcal{L}_{region} = 	\mathcal{L}_{dice}(\mathcal{S}(M, \widehat{\mathcal{P}}^{\sharp}), \ \mathcal{S}(\widehat{M}, \widehat{\mathcal{P}}^{\sharp}))
\end{align}
\vspace{-0.12cm}
where $\mathcal{L}_{dice}$ is the common dice loss function in \cite{milletari2016v}.
\\
\textit{\textbf{4)}}  \textit{overall \textit{\textbf{PCR-Loss}}.}
In order to exert the above three-level supervision at the same time, we put forward that the overall \textit{PCR-Loss } is a weighted sum of all three losses as, 
\vspace{-0.1cm}
\begin{align}
	\label{eq:pcr-loss}
	\mathcal{L}_{PCR} = \lambda_{p}\mathcal{L}_{point} + \lambda_{c}\mathcal{L}_{curve} + \lambda_{r}\mathcal{L}_{region}
\end{align}
\vspace{-0.6cm}

\minisection{Multi-task Auxiliary Loss.}
The paradigm of multi-task learning can reduce the risk of overfitting by leveraging the domain-specific information included in the training signals of related tasks \cite{zhang2021survey}.  
Thence, in addition to supervising the curve-level \Bezier modeling procedure of the final head, we also perform auxiliary segmentation-based supervision on the intermediate modules, \ie semantic-level \bev decoder and instance-level B\'ezier decoder. 
Given the output masks $\widehat{\mathbb{M}}_s$ and $\widehat{\mathbb{M}}_z$ in the Sec.\ref{sec:bemapnet}, we formulate the auxiliary loss with the combination of two tasks supervision as,
\vspace{-0.1cm}
\begin{align}
	\label{eq:aux-loss}
	\mathcal{L}_{AUX} = \lambda_{s}\mathcal{L}(\mathbb{M}_s, \widehat{\mathbb{M}}_s) + \lambda_{z}\mathcal{L}(\mathbb{M}_z, \widehat{\mathbb{M}}_z)
\end{align}
where $\mathbb{M}_{\star}$ denotes the ground truth and $\lambda_{\star}$ is the weighted factor. Note that $\mathcal{L}$ is a compound loss with common cross entropy loss and dice loss, namely $\mathcal{L} = \mathcal{L} _{ce}+ \mathcal{L}_{dice}$.

\section{Experiments}
\label{sec:experiment}

\begin{table*}[htb]
	\begin{center}
		\resizebox{1.0\textwidth}{!}{
			\begin{tabular}{lcc|cccc|ccccc}
				\hline
				\rowcolor{Gray}
				Method & Backbone & Epoch & AP$_{\textit{divider}}$ & AP$_{\textit{cross}}$ & AP$_{\textit{boundary}}$ & mAP & mAP$_{\textit{day}}$ & mAP$_{\textit{night}}$ & mAP$_{\textit{sunny}}$ & mAP$_{\textit{cloudy}}$ & mAP$_{\textit{rainy}}$  \\
				\toprule
				IPM(B) \cite{li2022hdmapnet} & Eff-B0 & 30 & 10.7 & \ \ 4.7 & 11.7 & \ \ 9.0 & - & - & - & - & - \\
				IPM(CB) \cite{li2022hdmapnet} & Eff-B0 & 30 & 24.0 & \ \ 7.3 & 27.8 & 19.7 & - & - & - & - & - \\
				LSS \cite{philion2020lift} & Eff-B0 & 30  & 22.9 & \ \ 5.1 & 24.2 & 17.4 & 11.5$^\flat$& 13.3$^\flat$ & 12.0$^\flat$ & 12.8$^\flat$ & \ \ 9.5$^\flat$ \\
				VPN \cite{pan2020cross} & Eff-B0 & 30 & 22.1 & \ \ 5.2 & 25.3 & 17.5 & 17.9$^\flat$& 17.5$^\flat$ & 18.2$^\flat$ & 18.5$^\flat$ & 15.9$^\flat$ \\
				HDMapNet \cite{li2022hdmapnet} & Eff-B0 & 30  & 28.3 & \ \ 7.1 & 32.6 & 22.7 & 21.4$^\flat$& 17.4$^\flat$ & 21.5$^\flat$ & 23.4$^\flat$ & 18.9$^\flat$ \\
				\midrule
				\textit{\model} & Eff-B0 & 30  & 46.7 & 37.4 & 38.0 & 40.7 & 41.3&  28.6& 41.4 & 44.9 & 36.2 \\
				\textit{\model} & Res-50 & 30  & 46.9 & 39.0 & 37.8 & 41.3 & 41.9 & 30.4 & 42.6 & 43.6 & 37.4 \\
				\bluecell{\textit{\model}} & \bluecell{Swin-T} & \bluecell{30}  & \bluecell{\textbf{49.1}} & \bluecell{\textbf{42.2}} & \bluecell{\textbf{39.9}} & \bluecell{\color{blue}\textbf{43.7}} & \bluecell{\textbf{44.3}}& \bluecell{\textbf{34.4}} &\bluecell{\textbf{44.8}} & \bluecell{\textbf{47.9}} & \bluecell{\textbf{37.8}}  \\
				\midrule
				\textit{\model} & Res-50 & 110  & 52.7 & 44.5 & 44.2 & 47.1 & 47.7 & 39.2 & 48.7 & 50.3 & 41.4 \\
				\textit{\model} & Swin-T & 110  & 54.2 & 46.5 & 46.5& 49.1 & 49.8& 37.3 & 50.4 & 53.5 & 42.8  \\
				\textit{\model} & Swin-B & 110  & 55.3 & 47.0 & 49.4 & 50.5 & 51.2 & 38.0 & 52.2 & 54.8 & 43.6  \\
				\midrule
				\midrule
				\greycell{\textit{\model}} & \greycell{Res-50} & \greycell30  & \greycell62.3 & \greycell57.7 & \greycell59.4 & \greycell59.8 & \greycell60.5 & \greycell46.9 & \greycell62.5 & \greycell61.9 & \greycell53.0 \\
				\greycell{\textit{\model}} & \greycell{Swin-T} & \greycell{30}  & \greycell{64.4} & \greycell61.3 & \greycell61.6& \greycell62.5 & \greycell63.1& \greycell53.2 & \greycell64.7 & \greycell66.0 & \greycell54.2  \\
				\greycell{\textit{\model}} & \greycell{Swin-B} & \greycell110  & \greycell69.0 & \greycell64.4 & \greycell69.7 & \greycell67.7 & \greycell68.3 & \greycell54.6 & \greycell70.3 & \greycell71.6 & \greycell58.0  \\
				\bottomrule
			\end{tabular}
		}
	\end{center}
	\vspace*{-0.5cm}
	\caption{
		Comparisons with \sota on \nuscene under thresholds of $[0.2, 0.5, 1.0]m$ and $[0.5, 1.0, 1.5]m$, where the results of latter easier evaluation protocol is marked by \colorbox[rgb]{0.9411,1.0,0.9411}{\textit{Green}} shade. 
		The \colorbox[rgb]{0.8549,0.9098,0.9882}{\textit{Blue}} shade contains the results that used in all ablation studies for a fair comparison.
		Note $^\flat$ and - indicate the results are re-implemented by us with public code and not available respectively.
	}
	\vspace*{-0.41cm}
	\label{tab:main-result-nuscenes}
\end{table*}

\subsection{Experimental Settings}
\minisection{Existing Benchmarks.}
To evaluate the proposed approach, we conduct experiments on the popular \nuscene dataset \cite{caesar2020nuscenes}, which consists of $28,130/6,019$ samples and $700/150$ driving scenes for the training/validation set respectively. 
Each scene contains roughly 40 samples and each sample includes 6 surrounding images, covering 360$^\circ$ FOV of the ego-vehicle.
For the sake of fair comparison, we follow the previous work \cite{li2022hdmapnet} and focus on three static map categories, \ie \LD, \PC and \RB.
Taking the ego-vehicle as the center, we set the perception range to $[30, 30, 15, 15]m$, which corresponds to the distances of front, rear, left and right respectively, and fix the resolution of ego-to-pixel as $0.15 \ m/pixel$.
In addition, for exploring the performance of our method under different lighting and weather conditions, we further divide \nuscene into five kinds of scene, namely \textit{day}, \textit{night}, \textit{sunny}, \textit{cloudy} and \textit{rainy}.
Refer to the supplementary material for more details.

\minisection{Evaluation Metrics.}
We utilize the exact same evaluation protocol as \cite{li2022hdmapnet} of average precision (AP) to access the map construction quality over the instance-level. 
To be concrete, given a pair of instances from ground-truth and predictions respectively, this protocol computes the Chamfer Distance between them and considers the prediction as true-positive only if the distance is less than a specified threshold, which is set to $[0.2, 0.5, 1.0]m$ in our experiment. Note the overall AP metric is obtained by averaging across three thresholds.
Moreover, as for a more informative comparison, the results under different lighting/weather conditions and a much simpler threshold setup of $[0.5, 1.0, 1.5]m$ are further provided.

\minisection{Implementation Details.}
We employ EfficientNet-B0 \cite{tan2019efficientnet}, ResNet-50 \cite{he2016deep} and SwinTR \cite{liu2021swin} as backbones, which are all initialized by ImageNet \cite{krizhevsky2017imagenet} pretraining. 
The following semantic \bev decoder stacks $2$ transformer encoder layers and $4$ decoder layers with $64 \times 32$ \ \bev queries, and the instance \Bezier decoder stacks $6$ mask-transformer decoder layers with $60$ queries, where $20, 25, 15$ for \LD, \PC and \RB respectively.
The shape of input image is resized to $896 \times 512$ and the mini-batch size is set to $1$ per GPU. 
We train our model with $8$ GPUs for $30/110$ epochs and adopt multi-step schedule with milestone $[0.7, 0.9]$ and $\gamma=\frac{1}{3}$.
The AdamW \cite{loshchilov2018decoupled} optimizer is employed with a weight decay of $1e^{-4}$ and a learning rate of $2e^{-4}$, which is multiplied by $0.1$ for backbone.
As for hyper-parameters of loss weight, we set $\lambda_{s}, \lambda_{z}, \lambda_{p}, \lambda_{c}, \lambda_{r}$ to $1, 5, 5, 10, 1$ respectively and the dilated width $\omega$ in $\mathcal{L}_{region}$ to $5$.
The deployments of piecewise \Bezier curve \bm{$\langle k, n \rangle$} for \LD, \PC and \RB are set to \bm{$\langle 3, 2 \rangle, \langle 1, 1 \rangle, \langle 7, 3 \rangle$}.
Note $m$ in Algorithm \ref{algo_pbc-generation} is set to $100$.

\subsection{Comparisons with State-of-the-art Methods}

\noindent
We present the overall evaluation results on \nuscene in Table \ref{tab:main-result-nuscenes}, which shows that our \modelbf is significantly superior to the existing \textit{SOTA} approaches by a large margin (up to \bm{$18.0$}) under the same setting of EfficientNet-B0 and $30$ epochs, indicating the effectiveness of our approach.
In addition, after replacing the backbone with more common ResNet-50 and more popular SwinTiny, our model achieves a further improvement of $0.6$ and $3.0$ AP respectively. 
Next, considering the slow convergence of the Transformer-based model, we increase the training schedule to $110$ epochs and gain at least another $5.4$ improvements.
Even in somewhat unconventional scenarios, such as night and rainy, our proposed approach still shows a great advantage, \ie $13.2$ and $16.9$ AP are obtained respectively.
Furthermore, another interesting observation is that regardless of which method is used, we find that the performance of sunny scene is always slightly lower than cloudy. We believe that strong illuminations and ground shades on sunny day might have a certain impact on the perception of map objects.
As a supplement, the last three rows of Table \ref{tab:main-result-nuscenes} show the performance of our approach on a much simpler evaluation protocol.

\begin{figure*}[htb]
	\begin{center}
		\includegraphics[width=0.9\linewidth]{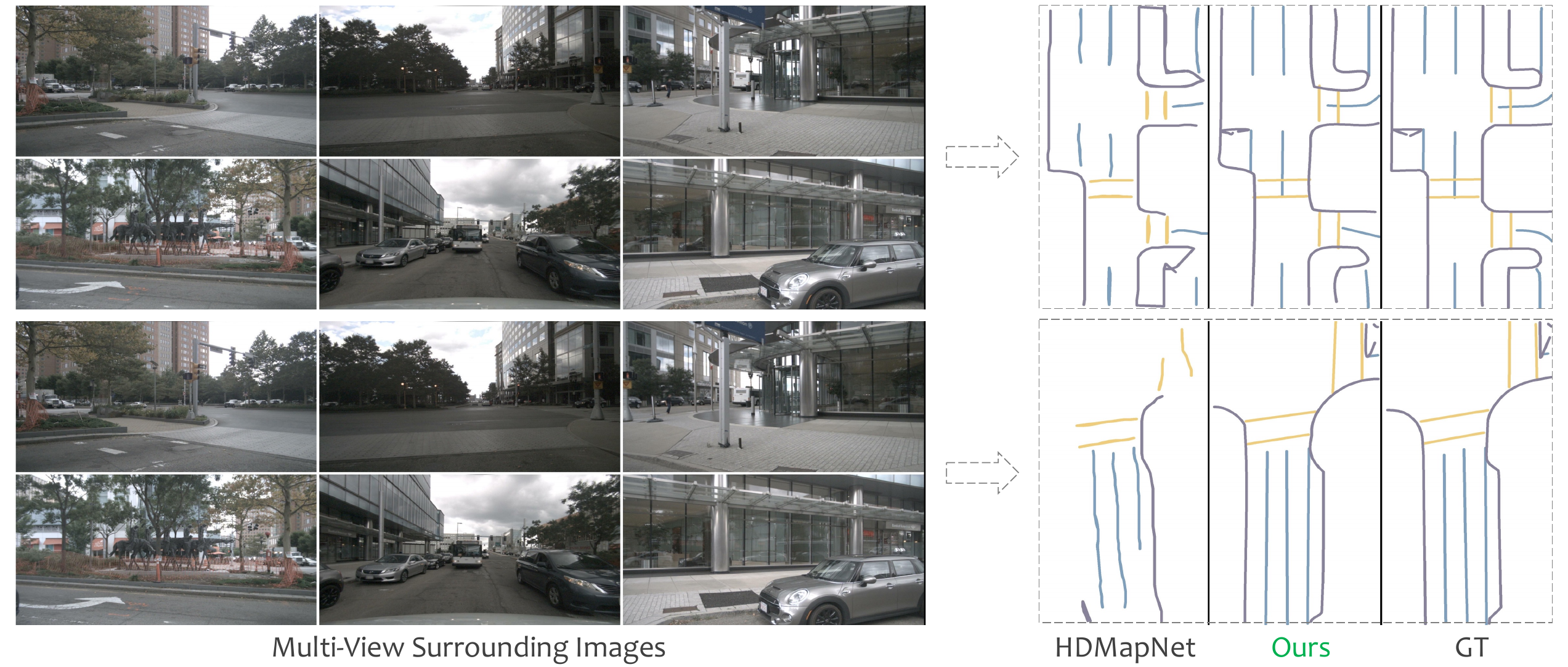}
	\end{center}
	\vspace{-0.57cm}
	\caption{
		The qualitative comparison results of \textit{HDMapNet} \cite{li2022hdmapnet}, \modelbf and \textit{GroundTruth} \cite{caesar2020nuscenes} under different scenarios.
	}
	\vspace{-0.44cm}
	\label{fig:vis_comparison}
\end{figure*} 

\subsection{Ablation Study}

\minisection{The Different Way of Position Encoding.}
Table \ref{tab:ablation-petype} shows the performance of different \textit{PE} in semantic \bev decoder for the map construction.
Compared with the \textit{sine/learned-based} method, our proposed \textit{IPM-PE} module improves AP by $3.9/5.7$ respectively, which proves the multi-perspective \textit{PE} guided by the intrinsic and extrinsic parameters is very effective.
Moreover, through the results in the $3$-$rd$ and $5$-$th$ row, we find that \textit{IPM-PE} with shared \textit{FC} layer alignment gains an further improvement of $2.6$ points.
Yet, in order to rule out the possibility of that the attached \textit{FC} layer brings an increment, we then feed the position embeddings from \textit{IPM-PE} into two \textbf{exclusive} \textit{FC}s and the $4$-$th$ row results exclude the influence of \textit{FC}.
These comparisons indicate the shared align layer is the exactly essential part to mitigate the performance decline brought by the unreasonable assumption of \textit{IPM}, which validates our conjecture in Sec.\ref{sec:ipmpe}.

\begin{table}[t]
	\begin{center}\centering
		\scalebox{0.87}{
			\begin{tabular}{c|ccc|c}
				\hline
				\rowcolor{Gray}
				Position Encoding & AP$_{\textit{divider}}$ & AP$_{\textit{cross}}$ & AP$_{\textit{boundary}}$ & mAP  \\
				\toprule
				Sine PE & 42.8 &34.1 & 34.7 & 37.2 \\
				Learned PE & 40.2 & 32.2 & 33.8 & 35.4 \\
				\midrule
				IPM-PE & 46.4 &  39.2 & 37.8 & 41.1 \\
				+ Exclusive \textit{FC}& 38.6 & 33.3 & 33.9 & 35.3  \\
				\bluecell{+ \textbf{Shared} \ \ \ \textit{FC}} & \bluecell{\textbf{49.1}} & \bluecell{\textbf{42.2}} & \bluecell{\textbf{39.9}} & \bluecell{\textbf{43.7}}  \\
				\bottomrule
			\end{tabular}
		}
	\end{center}
	\vspace*{-0.45cm}
	\caption{Comparisons on different methods of position encoding. }
	\vspace{-0.45cm}
	\label{tab:ablation-petype}
\end{table}

\minisection{Effectiveness of \textit{PCR-Loss}.}
The \textit{PCR-Loss} mainly consists of three parts, \ie point, curve and region. Table \ref{tab:ablation-recovery-loss} shows detailed ablation results of the role of each part.
Specifically, with adopting only one supervision in rows $1 \sim 3$, the performance trend is roughly: \textit{curve} $>$ \textit{point} $>$ \textit{region}.
Among them, it is worth noting that the only \textit{region-based} supervision shows very poor performance, which is understandable because the generation of region mask relies on a collection of relatively accurate points. This conjecture is also verified in rows $4 \sim 5$ of the Table \ref{tab:ablation-recovery-loss}.
Furthermore, when integrating multiple loss supervision, the final performance is consistently improved compared to the single one, and three parts used together achieves the optimal AP results of $43.7$.
\vspace{-0.1cm}
\begin{table}[ht]
	\begin{center}\centering
		\scalebox{0.84}{
			\begin{tabular}{ccc|ccc|c}
				\hline
				\rowcolor{Gray}
				Point & Curve & Region &AP$_{\textit{divider}}$ & AP$_{\textit{ped}}$ & AP$_{\textit{boundary}}$ & mAP  \\
				\toprule
				\cmark & \xmark & \xmark&  39.1& 33.3 & 23.7 & 32.0 \\
				\xmark & \cmark & \xmark & 47.2 & 35.1 & 35.9 &  39.4\\
				\xmark & \xmark& \cmark & \ \ 3.2 & \ \ 2.6 & \ \ 0.5 & \ \ 2.1 \\
				\cmark & \xmark & \cmark&  40.7&  39.4& 16.7 & 32.9 \\
				\xmark & \cmark & \cmark &  48.2&  38.5& 39.7 & 42.1 \\
				\cmark & \cmark & \xmark&  46.5& 38.5 & 36.2 & 40.4 \\
				\bluecell{\cmark} & \bluecell{\cmark}& \bluecell{\cmark} & \bluecell{\textbf{49.1}} & \bluecell{\textbf{42.2}} & \bluecell{\textbf{39.9}} & \bluecell{\textbf{43.7}}  \\
				\bottomrule
			\end{tabular}
		}
	\end{center}
	\vspace{-0.55cm}
	\caption{The effectiveness of different modules in \textit{PCR-Loss}.}
	\vspace{-0.35cm}
	\label{tab:ablation-recovery-loss}
\end{table}

\minisection{Effectiveness of Multi-task Auxiliary Loss.}
Without any auxiliary loss,  the average AP of our proposed \modelbf is only $20.0$.
Yet, after adopting another semantic/instance supervision, the result is improved to $34.5$/$38.4$ respectively and goes further to $43.7$ with employing both, which shows the auxiliary tasks for producing high-quality semantic \bev features and instance descriptors  are very important.
\vspace{-0.3cm}
\begin{table}[ht]
	\begin{center}\centering
		\scalebox{0.9}{
			\begin{tabular}{cc|ccc|c}
				\hline
				\rowcolor{Gray}
				Semantic & Instance &AP$_{\textit{divider}}$ & AP$_{\textit{ped}}$ & AP$_{\textit{boundary}}$ & mAP  \\
				\toprule
				\xmark & \xmark & 21.5 & 17.8 & 20.7 & 20.0 \\
				\cmark & \xmark&  39.3&  31.5& 32.8 & 34.5 \\
				\xmark & \cmark&  44.3& 34.6 & 36.5 & 38.4 \\
				\bluecell{\cmark} & \bluecell{\cmark}& \bluecell{\textbf{49.1}} & \bluecell{\textbf{42.2}} & \bluecell{\textbf{39.9}} & \bluecell{\textbf{43.7}}  \\
				\bottomrule
			\end{tabular}
		}
	\end{center}
	\vspace{-0.55cm}
	\caption{The effectiveness of different modules in \textit{AUX-Loss}.}
	\vspace{-0.25cm}
	\label{tab:ablation-aux-loss}
\end{table}

\minisection{Comparison with Polyline Vectorization.}
An amusing fact is that a \Bezier curve with $\bm{n}=1$ is simply a straight line between two control points.
By setting the degree of all elements to $1$, our framework naturally degenerates into the \textit{polyline-based} method.
Note we further set \bm{$k$} to $9, 1, 29$ for \LD, \PC and \RB respectively.
The comparisons in Table \ref{tab:ablation-paravspoly} shows the \Bezier-based vectorization approach achieves better $4.5$ AP.
Interestingly, as the complexity of map element shape increases (usually \PC $<$ \LD $<$ \RB), we find that the corresponding \textit{AP} improvement increases as well.

\vspace{-0.2cm}
\begin{table}[ht]
	\begin{center}\centering
		\scalebox{0.89}{
			\begin{tabular}{c|ccc|c}
				\hline
				\rowcolor{Gray}
				Vectorization & AP$_{\textit{divider}}$ & AP$_{\textit{ped}}$ & AP$_{\textit{boundary}}$ & mAP  \\
				\toprule
				Polyline & 45.0 & 39.1 & 33.4 & 39.2 \\
				\bluecell{\Bezier} & \bluecell{\textbf{49.1} \textbf{\scriptsize{\color{blue}(+4.1)}}} & \bluecell{\textbf{42.2} \textbf{\scriptsize{\color{blue}(+3.1)}}} & \bluecell{\textbf{39.9} \textbf{\scriptsize{\color{blue}(+6.5)}}}& \bluecell{\textbf{43.7}\textbf{\scriptsize{\color{blue}(+4.5)}}}  \\
				\bottomrule
			\end{tabular}
		}
	\end{center}
	\vspace{-0.55cm}
	\caption{Comparison with the different type of vectorization. }
	\vspace{-0.3cm}
	\label{tab:ablation-paravspoly}
\end{table}

\minisection{Qualitative Analysis.}
We show the qualitative comparisons with \sota in Fig. \ref{fig:vis_comparison}.  
Besides avoiding complex vectorized post-processing, the proposed \textit{\model} constructs various and changeful \textit{HD-}elements more compactly and robust. 
Note that more extensive visualizations under different conditions are further provided in our supplementary material.

\section{Conclusion}
Vectorized \hdmap online construction focuses on the perception of centimeter-level environmental information. Starting from the conventional parameterization-based methods, this paper presents an end-to-end postprocessing-free architecture, namely \modelbf, with leveraging unified piecewise \Bezier curve for various and changeful map elements. By introducing three well-designed modules, \ie \textit{IPM-PE Align Module}, \textit{Piecewise \Bezier Head} and \textit{Point-Curve-Region Loss}, the overall framework is concise and elegant, which reaches state-of-the-art performance and provides a new perspective for future \hdmap research.

\clearpage

\appendix

\section*{\centering Supplementary Material}
In this supplementary material, we provide additional details which we could not include in the main paper due to space limitations, including more experimental analysis and visualization details that help us develop further insights to the proposed \modelbf. We discuss:

\begin{itemize}
	\setlength\itemsep{-0.15em}
	\item Additional details of architecture design.
	\item More experimental results for map construction.
	\item The statistical analysis of benchmark extension.
	\item Qualitative visualization results of our approach.
\end{itemize}

\section{Additional Details of Architecture Design}

\subsection{The Motivation of IPM-PE}
\begin{figure}[htb]
	\begin{center}
		\includegraphics[width=0.96\linewidth]{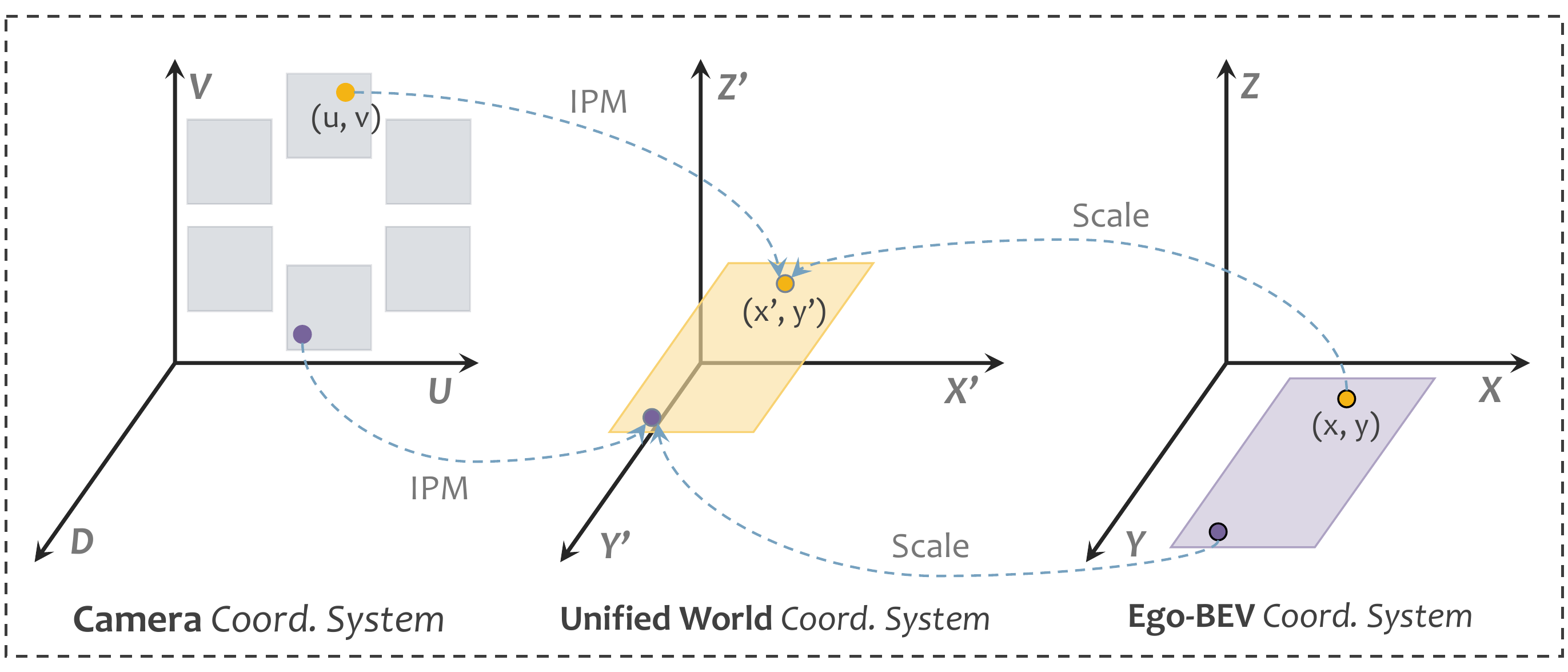}
	\end{center}
	\vspace{-0.57cm}
	\caption{Illustration of the motivation of IPM-PE.}
	\vspace{-0.44cm}
	\label{fig:ipmpe_motivation}
\end{figure}

According to the principle of perspective geometry, a pair of corresponding points in the camera space and \bev space is theoretically the projection result from the same point in the unified world coordinate system.
Based on this prior, on the one hand, we perform IPM to map multi-view camera coordinate grids into the world coordinate system, on the other hand, the \bev feature grid is also transformed into the same world space through a scale coefficient. 
This process is illustrated in detail in Fig.\ref{fig:ipmpe_motivation}.
Compared with the conventional positional encoding method that establishes the correspondence within each single-view, the proposed IPM-PE models the position relationship of \textit{\textbf{multi-view}} and \textit{\textbf{multi-space}} simultaneously, which is more conducive to perspective transformation.
Table \ref{tab:ablation-ipmpe-motivation} shows the effectiveness of the proposed IPM-PE module.
\vspace{-0.2cm}

\vspace{-0.1cm}
\begin{table}[ht]
	\begin{center}\centering
		\scalebox{0.92}{
			\begin{tabular}{l|ccc|c}
				\hline
				\rowcolor{Gray}
				PE Type & AP$_{\textit{divider}}$ & AP$_{\textit{ped}}$ & AP$_{\textit{boundary}}$ & mAP  \\
				\toprule
				Learned PE & 40.2 & 32.2 & 33.8 & 35.4 \\
				IPM-PE & \textbf{46.4} & \textbf{39.2} & \textbf{37.8} & \textbf{41.1} \\
				\bottomrule
			\end{tabular}
		}
	\end{center}
	\vspace{-0.55cm}
	\caption{The impact of different number of \bev queries. }
	\vspace{-0.4cm}
	\label{tab:ablation-ipmpe-motivation}
\end{table}

\subsection{Alignment Method of IPM-PE}
\begin{figure}[htb]
	\begin{center}
		\includegraphics[width=0.96\linewidth]{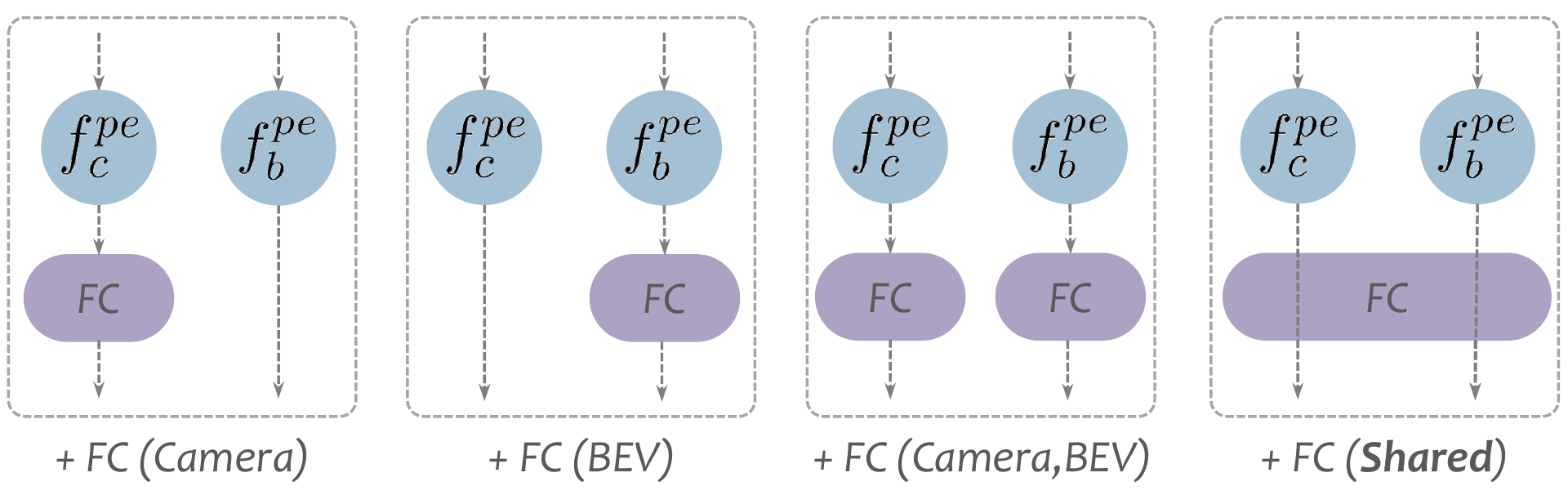}
	\end{center}
	\vspace{-0.57cm}
	\caption{
		Illustration of different align methods of IPM-PE.
	}
	\vspace{-0.44cm}
	\label{fig:ipmpe_more}
\end{figure}

We illustrate four align methods with common \textit{FC} layer in Fig.\ref{fig:ipmpe_more}, namely camera-only, \bev-only, \textit{exclusive} camera-\bev, and \textit{shared} camera-\bev.
Table \ref{tab:ablation-ipmpe-more} summarizes the experimental results in detail.
The comparison of row $1$ and $2\sim4$ indicates that the single or exclusive alignment instead causes performance decline. 
We conjecture that none of these three align methods share any information between the two branches, and the additional \textit{FC} instead arouse positional embedding from IPM-PE to lose the original accurate geometric prior.
Interestingly, with adopting the setting of shared camera-\bev, the model gains performance improvement with $2.6$ AP.
We argue that this is due to these shared parameters bridge the two groups of positional embeddings (\ie $f^{pe}_c$ and $f^{pe}_b$) with mutual alignment, which neutralizes the unreasonable assumptions in IPM to a certain extent.

\vspace{-0.1cm}
\begin{table}[ht]
	\begin{center}\centering
		\scalebox{0.87}{
			\begin{tabular}{l|ccc|c}
				\hline
				\rowcolor{Gray}
				Align Type & AP$_{\textit{divider}}$ & AP$_{\textit{ped}}$ & AP$_{\textit{boundary}}$ & mAP  \\
				\toprule
				IPM-PE& 46.4 & 39.2 & 37.8 & 41.1 \\
				\midrule
				\midrule
				$+$ FC (Camera)&  42.1 & 34.1 & 36.3 & 37.5 \\
				$+$ FC (\bev)& 46.9 & 38.6 & 35.9 & 40.5 \\
				$+$ FC (Camera,\bev)& 38.6 & 33.3 & 33.9&35.3 \\
				{$+$ FC (Shared)}& {\textbf{49.1}} & {\textbf{42.2}} & {\textbf{39.9}} & {\textbf{43.7}} \\
				\bottomrule
			\end{tabular}
		}
	\end{center}
	\vspace{-0.55cm}
	\caption{The impact of different align methods of IPM-PE. }
	\vspace{-0.4cm}
	\label{tab:ablation-ipmpe-more}
\end{table}

\subsection{Coordinate Regression Head Design}
\begin{figure}[htb]
	\begin{center}
		\includegraphics[width=0.96\linewidth]{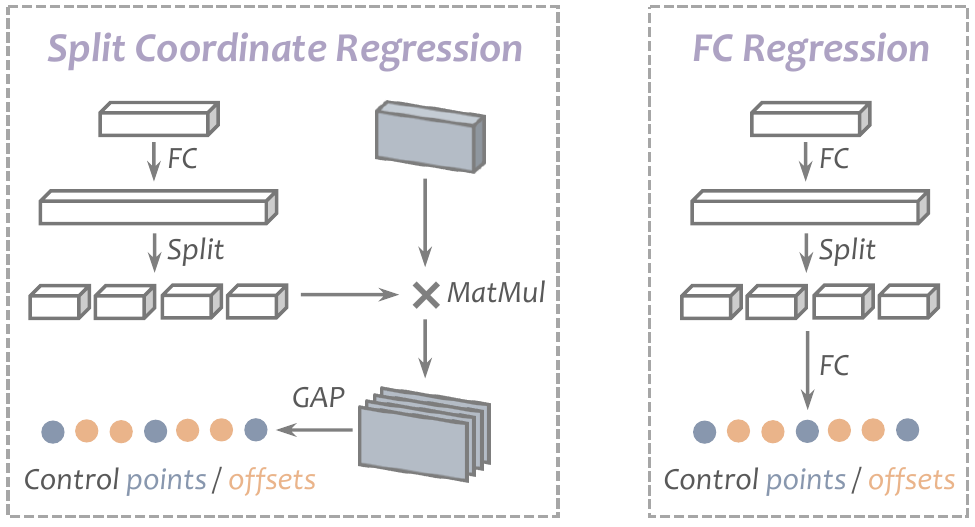}
	\end{center}
	\vspace{-0.57cm}
	\caption{Different design of coordinate regression head.}
	\vspace{-0.44cm}
	\label{fig:output_head}
\end{figure}

Inspired by the popular dynamic conv \cite{chen2020dynamic}, we propose the Split Coordinate Regression Head in the main paper and illustrate its detailed framework in the Fig.\ref{fig:output_head} (left).
In fact, there is a more direct and common point regression method, which directly predicts the  coordinate value through conducting a \textit{FC} projection on the splitted \Bezier descriptor, as shown in the Fig.\ref{fig:output_head} (right).
We compared the performance of these two approaches in Table \ref{tab:ablation-outputhead}, which demonstrates that the proposed dynamic way is more advantageous.

\vspace{-0.1cm}
\begin{table}[ht]
	\begin{center}\centering
		\scalebox{0.86}{
			\begin{tabular}{c|ccc|c}
				\hline
				\rowcolor{Gray}
				Shape & AP$_{\textit{divider}}$ & AP$_{\textit{ped}}$ & AP$_{\textit{boundary}}$ & mAP  \\
				\toprule
				\textit{FC Reg.} &  48.7 & 38.9 & 38.7 & 42.1 \\
				{\textit{Split Coord. Reg.}}& {\textbf{49.1}} & {\textbf{42.2}} & {\textbf{39.9}} & {\textbf{43.7}} \\
				\bottomrule
			\end{tabular}
		}
	\end{center}
	\vspace{-0.55cm}
	\caption{The impact of different methods of regression head. }
	\vspace{-0.4cm}
	\label{tab:ablation-outputhead}
\end{table}

\section{More Experimental Results }

\subsection{The running time and model size.}
The Table \ref{tab:speed} below provides more detailed performance comparison between \textit{HDMapNet} and \modelbf, including inference speed and parameter quantity. 
Note all numbers are obtained on one RTX $2080$Ti GPU with $bs=1$.
Under the same backbone setting of EfficientNet-B0 \cite{tan2019efficientnet}, our approach achieves better results with $0.4 \times$ fewer parameters ($69.8 \rightarrow 27.6$) and $6.9 \times$ higher speed ($0.7 \rightarrow 4.8$). 
Note that the inference time is the total time of network forwarding and post-processing, excluding the data loading time. The \textit{FPS} is obtained averagely on the whole validation set.
\vspace{-0.15cm}
\begin{table}[ht]
	\begin{center}\centering
		\scalebox{0.9}{
			\begin{tabular}{c|c|ccc}
				\hline
				\rowcolor{Gray}
				Method & Backbone & mAP $\uparrow$ & FPS $\uparrow$ & Params (MB) $\downarrow$ \\
				\toprule
				\textit{HDMapNet} & Eff-B0 & 22.7 & 0.7 & 69.8 \\
				\midrule
				\textit{\model} & Eff-B0 & 40.7 & \textbf{4.8} & \textbf{27.6}  \\
				\textit{\model} & SwinT & \textbf{43.7} & 3.7 & 55.3  \\
				\bottomrule
			\end{tabular}
		}
	\end{center}
	\vspace{-0.55cm}
	\caption{The comparison of running time and model size.}
	\label{tab:speed}
	\vspace{-0.4cm}
\end{table}

\subsection{The Number of Transformer Layers.}
Our \modelbf is a DETR-like architecture and we explore the influence of the number of transformer layers in Table \ref{tab:ablation-num-enc-dec}.
As the number of layers increases, the overall AP improves gradually, but the performance tends to be saturated when the number reaches $6$.
Note that when we reduce the number of layers to $1$ for all three modules, the performance can still reach $30.9$ \textit{AP}.
\vspace{-0.1cm}
\begin{table}[ht]
	\begin{center}\centering
		\scalebox{0.9}{
			\begin{tabular}{c|ccc|c}
				\hline
				\rowcolor{Gray}
				$(N^{bev}_{enc}, N^{bev}_{dec}, N^{ins}_{dec})$ & AP$_{\textit{divider}}$ & AP$_{\textit{ped}}$ & AP$_{\textit{boundary}}$ & mAP  \\
				\toprule
				$[\ 1, \ 1, \ 1\ ]$ & 35.4 & 27.8 & 29.5 & 30.9 \\
				$[\ 2, \ 2, \ 2\ ]$ & 44.1 & 36.0 & 36.5 & 38.9 \\
				$[\ 2, \ 2, \ 6\ ]$ & 45.9 & 38.4 & 36.1& 40.1 \\
				$[\ 2, \ 4, \ 6\ ]$ & 49.1 & 42.2 & 39.9 &43.7  \\
				$[\ 2, \ 6, \ 6\ ]$ & 50.8 & 43.0 & 41.8 & 45.2  \\
				$[\ 6, \ 4, \ 6\ ]$ & \textbf{52.1} & \textbf{44.2} & \textbf{43.0} & \textbf{46.4}  \\
				$[\ 6, \ 6, \ 6\ ]$ & 51.9 & 42.7 & 42.9 &45.8  \\
				\bottomrule
			\end{tabular}
		}
	\end{center}
	\vspace{-0.55cm}
	\caption{The impact of different number of transformer layers. }
	\vspace{-0.4cm}
	\label{tab:ablation-num-enc-dec}
\end{table}

\subsection{The Number of \bev Queries.}
According to the main paper, each query models the feature of a specific region on \bev.
In other words, the number of queries represents the resolution of \bev feature, which is closely related to the final AP performance. Table \ref{tab:ablation-bevqueryshape} verifies the above conjecture.
\vspace{-0.1cm}
\begin{table}[ht]
	\begin{center}\centering
		\scalebox{0.96}{
			\begin{tabular}{c|ccc|c}
				\hline
				\rowcolor{Gray}
				$\#$ \bev Queries & AP$_{\textit{divider}}$ & AP$_{\textit{ped}}$ & AP$_{\textit{boundary}}$ & mAP  \\
				\toprule
				$\ \ 8 \ \times \ \ 16$ & 39.3 & 31.9 & 33.3&34.8 \\
				$16 \ \times \ \ 32$ & 44.6 & 39.3 & 37.3 & 40.4 \\
				$32 \ \times \ \ 64$ & 49.1 & 42.2 & 39.9 & 43.7 \\
				$64 \ \times 128$ & \textbf{51.3} & \textbf{43.5} & \textbf{42.1} & \textbf{45.6}  \\
				$80 \ \times 160$ & 51.1 & 43.2 & 39.6 &44.6  \\
				\bottomrule
			\end{tabular}
		}
	\end{center}
	\vspace{-0.55cm}
	\caption{The impact of different number of \bev queries. }
	\vspace{-0.4cm}
	\label{tab:ablation-bevqueryshape}
\end{table}

\subsection{The Number of \Bezier Queries.}
The instance \Bezier decoder infers curve predictions with a fixed-size set, which is usually larger than the typical number of map elements. 
Table \ref{tab:ablation-bezierqueries} shows the impact with different number of queries for \LD, \PC, \RB.
\vspace{-0.1cm}
\begin{table}[ht]
	\begin{center}\centering
		\scalebox{0.94}{
			\begin{tabular}{c|ccc|c}
				\hline
				\rowcolor{Gray}
				$\#$ \Bezier Queries & AP$_{\textit{divider}}$ & AP$_{\textit{ped}}$ & AP$_{\textit{boundary}}$ & mAP  \\
				\toprule
				$[\ 10, \ 12, \ \ \ 8 \ ]$ & 47.7 & 39.5 & 39.1 & 42.1  \\
				$[\ 20, \ 25, \ 15 \ ]$ & 49.1 & \textbf{42.2} & 39.9 & 43.7 \\
				$[\ 30, \ 36, \ 24 \ ]$ & \textbf{50.2} & 41.4 & \textbf{40.0} & \textbf{43.8}  \\
				\bottomrule
			\end{tabular}
		}
	\end{center}
	\vspace{-0.55cm}
	\caption{The impact of different number of \Bezier queries. }
	\vspace{-0.4cm}
	\label{tab:ablation-bezierqueries}
\end{table}

\subsection{The Impact of Different \Bezier Setup}
For piecewise \Bezier curve, various \bm{$n$} and \bm{$k$} determine different fitting capabilities. Taking \RB as an example, we explore the impact in different \Bezier setup on model performance from three aspects as follows,
\\ \noindent 1) fixed \bm{$k$}. 
The curve with larger $n$ has stronger curve fitting ability, but it will also increase the learning difficulty of the model. The results in Table \ref{tab:ablation-diff-degree} confirm this conjecture.
\\ \noindent 2) fixed  \bm{$n$}. 
As shown in Table \ref{tab:ablation-diff-piece}, as the piece number gradually increases, the model performance gradually improves and tends to be saturated when $k$ reaches a certain value. 
\\ \noindent 3) fixed the number of control points \bm{$nk+1$}. 
With exactly same expressive ability, Table \ref{tab:ablation-diff-degree-diff-piece} proves that too many pieces and too large degree both degrade the performance.

\vspace{-0.1cm}
\begin{table}[ht]
	\begin{center}\centering
		\scalebox{0.97}{
			\begin{tabular}{c|c}
				\hline
				\rowcolor{Gray}
				\bm{$\langle k, n \rangle$} & AP$_{\textit{boundary}}$  \\
				\toprule
				$k=8, \ n=1$& 33.4 \\
				$k=8, \ n=2$& 39.2  \\
				$k=8, \ n=3$& \textbf{39.6} \\
				$k=8, \ n=4$& 39.5\\
				\bottomrule
			\end{tabular}
		}
	\end{center}
	\vspace{-0.55cm}
	\caption{The impact of different number of degree $n$ ($k=8$). }
	\vspace{-0.4cm}
	\label{tab:ablation-diff-degree}
\end{table}

\vspace{-0.1cm}
\begin{table}[ht]
	\begin{center}\centering
		\scalebox{0.97}{
			\begin{tabular}{c|c}
				\hline
				\rowcolor{Gray}
				\bm{$\langle k, n \rangle$}  & AP$_{\textit{boundary}}$  \\
				\toprule
				$k=\ \ 2, \ n=3$& 32.6 \\
				$k=\ \ 4, \ n=3$& 38.4 \\
				$k=\ \ 6, \ n=3$& 39.1 \\
				$k=\ \ 7, \ n=3$&\textbf{39.9}  \\
				$k=\ \ 8, \ n=3$& 39.6 \\
				$k=10, \ n=3$& 39.4  \\
				$k=12, \ n=3$& 39.6 \\
				\bottomrule
			\end{tabular}
		}
	\end{center}
	\vspace{-0.55cm}
	\caption{The impact of different number of piece $k$  ($n=3$).}
	\vspace{-0.4cm}
	\label{tab:ablation-diff-piece}
\end{table}

\vspace{-0.1cm}
\begin{table}[ht]
	\begin{center}\centering
		\scalebox{0.97}{
			\begin{tabular}{c|c}
				\hline
				\rowcolor{Gray}
				\bm{$\langle k, n \rangle$}  & AP$_{\textit{boundary}}$\\
				\toprule
				$k=24, \ n=1$& 33.8   \\
				$k=12, \ n=2$& 37.7  \\
				$k=\ \ 8, \ n=3$& \textbf{39.6}  \\
				$k=\ \ 6, \ n=4$& 39.1 \\
				\bottomrule
			\end{tabular}
		}
	\end{center}
	\vspace{-0.55cm}
	\caption{The impact of different number of $n$ and $k$ ($nk=24$). }
	\vspace{-0.4cm}
	\label{tab:ablation-diff-degree-diff-piece}
\end{table}

\subsection{The Impact of Input Resolution}
Comparing the $1$-\textit{st} and $4$-\textit{th} row in Table \ref{tab:ablation-inputshape}, the resolution is increased by $8.4\times$, and the performance is improved by $4.5$ AP. We adopt resolution $896\times512$ in all experiments.

\vspace{-0.1cm}
\begin{table}[ht]
	\begin{center}\centering
		\scalebox{0.95}{
			\begin{tabular}{c|ccc|c}
				\hline
				\rowcolor{Gray}
				Input Size & AP$_{\textit{divider}}$ & AP$_{\textit{ped}}$ & AP$_{\textit{boundary}}$ & mAP  \\
				\toprule
				$\ \ 512 \times 320$ & 46.2 & 37.5 & 38.7 & 40.8  \\
				$\ \ 640 \times 384$& 47.9 & 38.9 & 38.5 & 41.8  \\
				$\ \ 896 \times 512$&49.1 & 42.2 & 39.9 & 43.7  \\
				$1536 \times 896$& \textbf{50.4} & \textbf{43.3} & \textbf{42.4} & \textbf{45.3}  \\
				\bottomrule
			\end{tabular}
		}
	\end{center}
	\vspace{-0.55cm}
	\caption{The impact of different input resolution. }
	\vspace{-0.4cm}
	\label{tab:ablation-inputshape}
\end{table}

\subsection{The Impact of Different FPN-Output Shape}
We directly interpolate the multi-scale features from FPN to a fixed size and then concat them together as the input of the subsequent \bev decoder.
As the upsample shape increases, the overall AP improves gradually, but the performance tends to be saturated when the shape reaches $30 \times 70$.

\vspace{-0.1cm}
\begin{table}[ht]
	\begin{center}\centering
		\scalebox{0.95}{
			\begin{tabular}{c|ccc|c}
				\hline
				\rowcolor{Gray}
				UpSample Shape & AP$_{\textit{divider}}$ & AP$_{\textit{ped}}$ & AP$_{\textit{boundary}}$ & mAP  \\
				\toprule
				$12 \times 28$& 47.8 & 38.4 & 39.2&41.8 \\
				$21 \times 49$& 49.1 & 42.2 & 39.9 & 43.7 \\
				$30 \times 70$& \textbf{50.3} & \textbf{43.3} & \textbf{42.0} & \textbf{45.2}  \\
				$36 \times 84$& 50.2 & 42.7 & 41.6 & 44.8  \\
				\bottomrule
			\end{tabular}
		}
	\end{center}
	\vspace{-0.55cm}
	\caption{The impact of different output shape of FPN. }
	\vspace{-0.4cm}
	\label{tab:ablation-fpnshape}
\end{table}

\subsection{The Impact of Different Curve Length}

According to the matrix form of the \Bezier definition, $P=B \times C$, where $B \in \mathbb{R}^{m \times n}$ and $m$ is the point number of the curve, \ie curve length, which 
is closely related to the curve supervision part in \textit{PCR-Loss}.
Table \ref{tab:ablation-pointnum} summarizes the performance with different curve length.

\vspace{-0.1cm}
\begin{table}[ht]
	\begin{center}\centering
		\scalebox{0.95}{
			\begin{tabular}{c|ccc|c}
				\hline
				\rowcolor{Gray}
				Curve Length & AP$_{\textit{divider}}$ & AP$_{\textit{ped}}$ & AP$_{\textit{boundary}}$ & mAP  \\
				\toprule
				$25$& 46.2 & 40.2 & 37.7 & 41.4  \\
				$50$& 48.0 & 39.8 & 39.3 & 42.4 \\
				$100$& \textbf{49.1} & \textbf{42.2} & \textbf{39.9} & \textbf{43.7}  \\
				$200$& 47.8 & 40.0 & 38.4 & 42.0 \\
				\bottomrule
			\end{tabular}
		}
	\end{center}
	\vspace{-0.55cm}
	\caption{The impact of different length of restored \Bezier curve. }
	\vspace{-0.4cm}
	\label{tab:ablation-pointnum}
\end{table}

\subsection{More details of the proposed \textit{\textbf{GenGT}}}
\noindent
Line-1$\&$2$\&$3: initialize the Bernstein coefficient matrix and its pseudo-inverse with given $n, m$ through math definition.
\\ \noindent
Line-4: keep the loop condition: \textit{start index} $<$ \textit{end index}.
\\ \noindent
Line-5: interpolate the $(s, e)$ sub-curve to length $m$(get $\mathit{P}^{\dagger}$).
\\ \noindent
Line-6: get control points $\mathbb{C}^{\dagger}$ of $\mathit{P}^{\dagger}$ by least squares fitting.
\\ \noindent
Line-7: restore its \Bezier curve $\mathit{P}^{\ddagger}$ accurately by above $\mathbb{C}^{\dagger}$.
\\ \noindent
Line-8: compute fitting error (\textit{CD-dist}) between $\mathit{P}^{\dagger}$ and $\mathit{P}^{\ddagger}$.
\\ \noindent
Line-9$\&$10$\&$11: update start/end index based on fit status.

\subsection{Verification of the Generated \Bezier GTs}
To verify the reliability of the ground truth produced by Algorithm \textit{\textbf{GTGen}} in the main paper, we first restore the generated control point sequences to their corresponding  \Bezier curves, and then treat these curves as predictions and the original annotations as ground-truths.
By conducting the exact same evaluation protocol, we calculate the AP performance between them, see Table \ref{tab:verify-on-gengt} for details.  When the Chamfer Distance threshold of true positive (termed as $\tau$ in Table \ref{tab:verify-on-gengt}) is set to $0.2m$, no matter  which degree of the curve is adopted, the overall mAP can reach more than $99.94$. 
Moreover, under the more tight setting of $\tau=0.1m$, each mAP can still achieve $97.7 \sim 98.7$.

\vspace{-0.1cm}
\begin{table}[ht]
	\begin{center}\centering
		\scalebox{0.95}{
			\begin{tabular}{c|c|ccc|c}
				\hline
				\rowcolor{Gray}
				Degree & $\tau$ &AP$_{\textit{divider}}$ & AP$_{\textit{ped}}$ & AP$_{\textit{boundary}}$ & mAP  \\
				\toprule
				$1$& 0.1 & 97.082 & 99.485 & 96.599 & 97.722 \\
				$2$& 0.1 &98.978 & 99.485 & 96.949 & 98.471 \\
				$3$& 0.1 &99.401 & 99.485 & 96.221 & 98.369 \\
				$4$& 0.1 &99.803 & 99.485 & 96.726 & 98.671 \\
				\midrule
				\midrule
				$1$& 0.2 &99.955 & 99.892 & 99.998 & 99.948 \\
				$2$& 0.2 &99.955 & 99.892 & 99.999 & 99.949\\
				$3$& 0.2 &99.955 & 99.892 & 99.995 & 99.947 \\
				$4$& 0.2 &99.955 & 99.892 & 99.999 & 99.949 \\
				\bottomrule
			\end{tabular}
		}
	\end{center}
	\vspace{-0.55cm}
	\caption{The reliability verification of generated \Bezier GTs. }
	\vspace{-0.4cm}
	\label{tab:verify-on-gengt}
\end{table}

\section{Statistical Analysis of Benchmark}

\subsection{Data Split by Different Conditions}
\begin{figure}[htb]
	\begin{center}
		\includegraphics[width=1.0\linewidth]{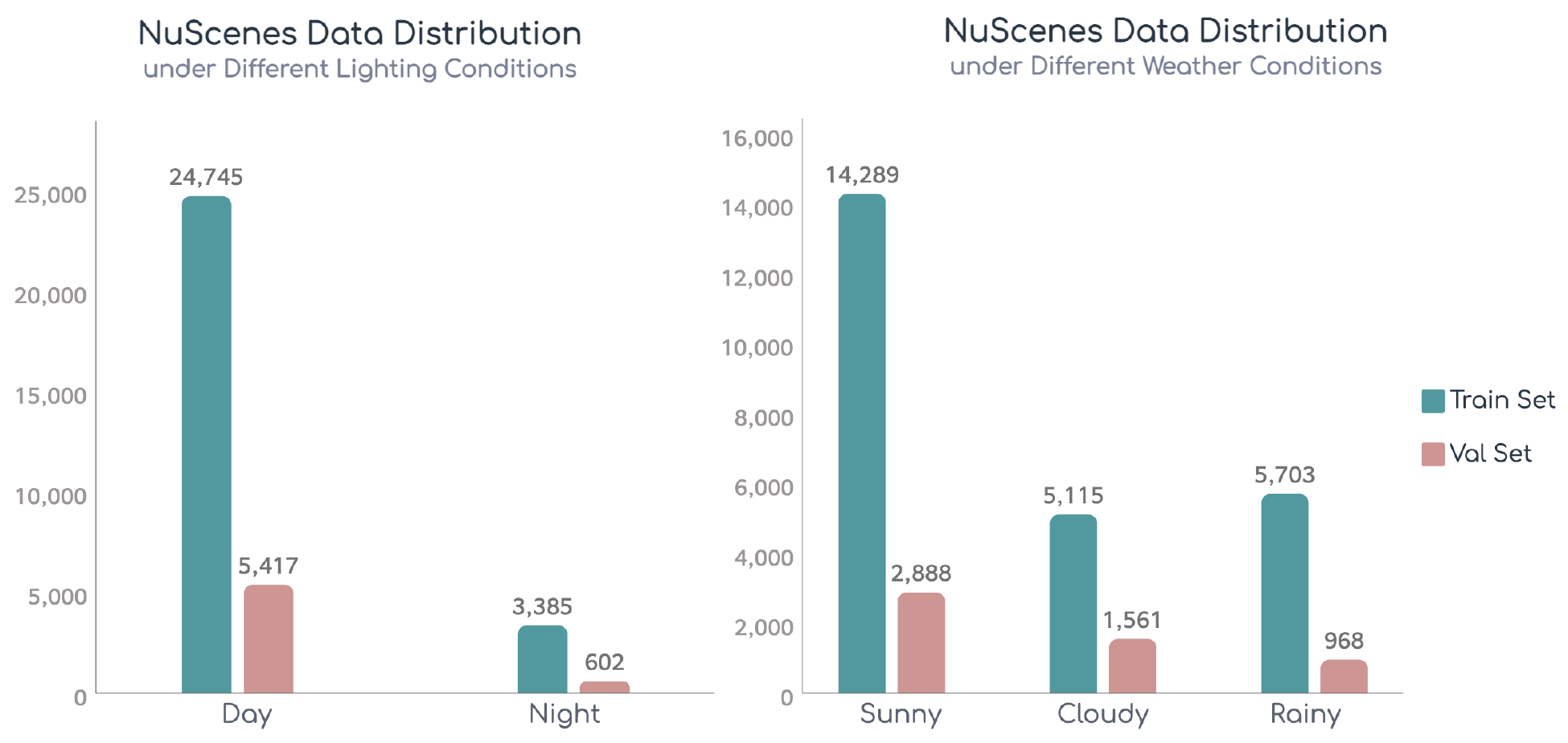}
	\end{center}
	\vspace{-0.57cm}
	\caption{Data distribution under different conditions.}
	\vspace{-0.44cm}
	\label{fig:sat_condition}
\end{figure}

\begin{figure}[htb]
	\begin{center}
		\includegraphics[width=0.88\linewidth]{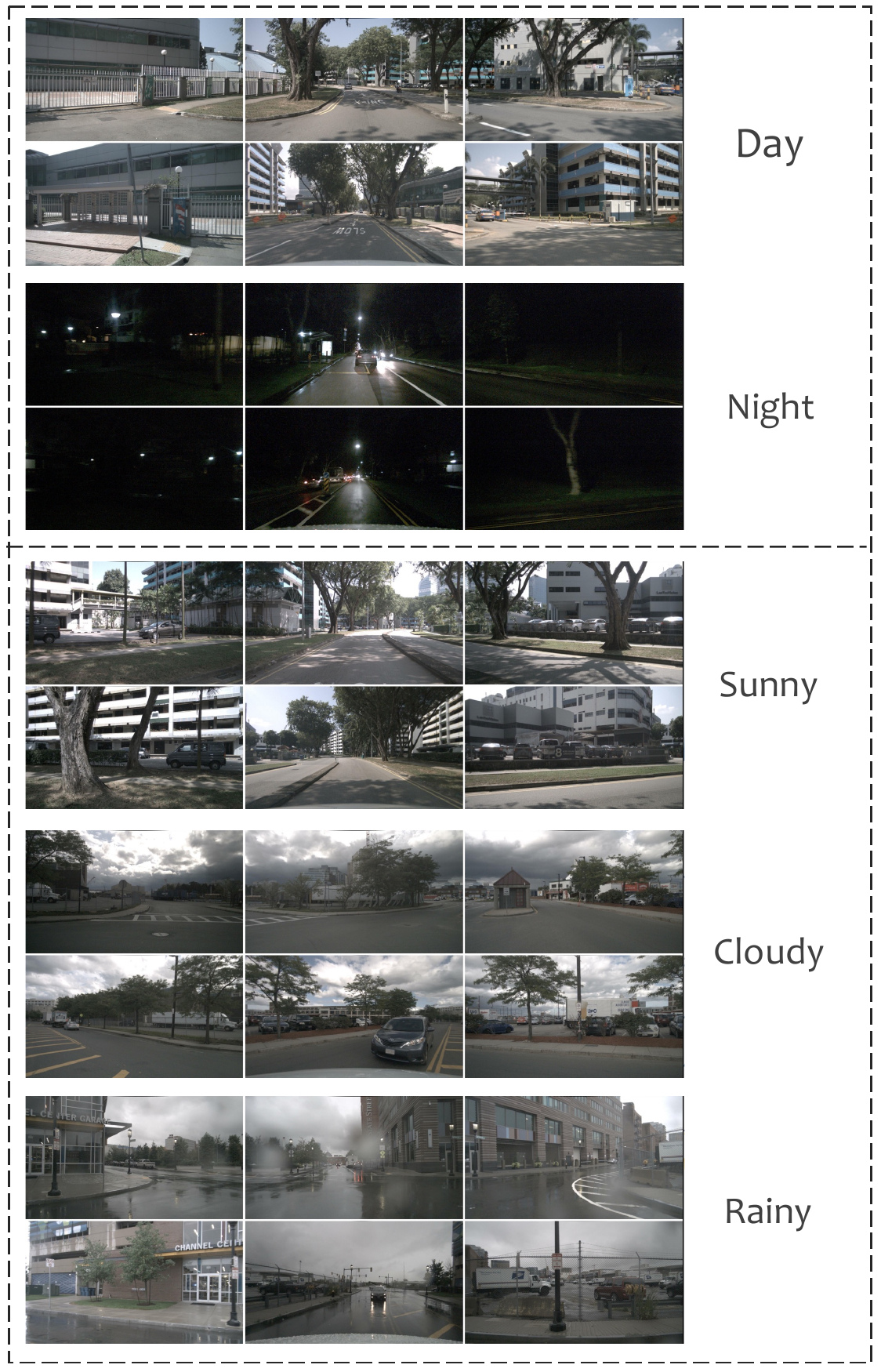}
	\end{center}
	\vspace{-0.57cm}
	\caption{Scenario examples under different conditions.}
	\vspace{-0.44cm}
	\label{fig:vis_condition}
\end{figure}

In order to explore the effectiveness of \modelbf under different lighting and weather conditions, we further divide \nuscene \cite{caesar2020nuscenes} into five kinds of scene, \ie \textit{day}, \textit{night}, \textit{sunny}, \textit{cloudy}, and \textit{rainy}.
Fig.\ref{fig:sat_condition}$\sim$\ref{fig:vis_condition}  provide the data distribution and scenario example for different subsets respectively.

\subsection{More Compact Map Element Expression}
\begin{figure}[htb]
	\begin{center}
		\includegraphics[width=1.0\linewidth]{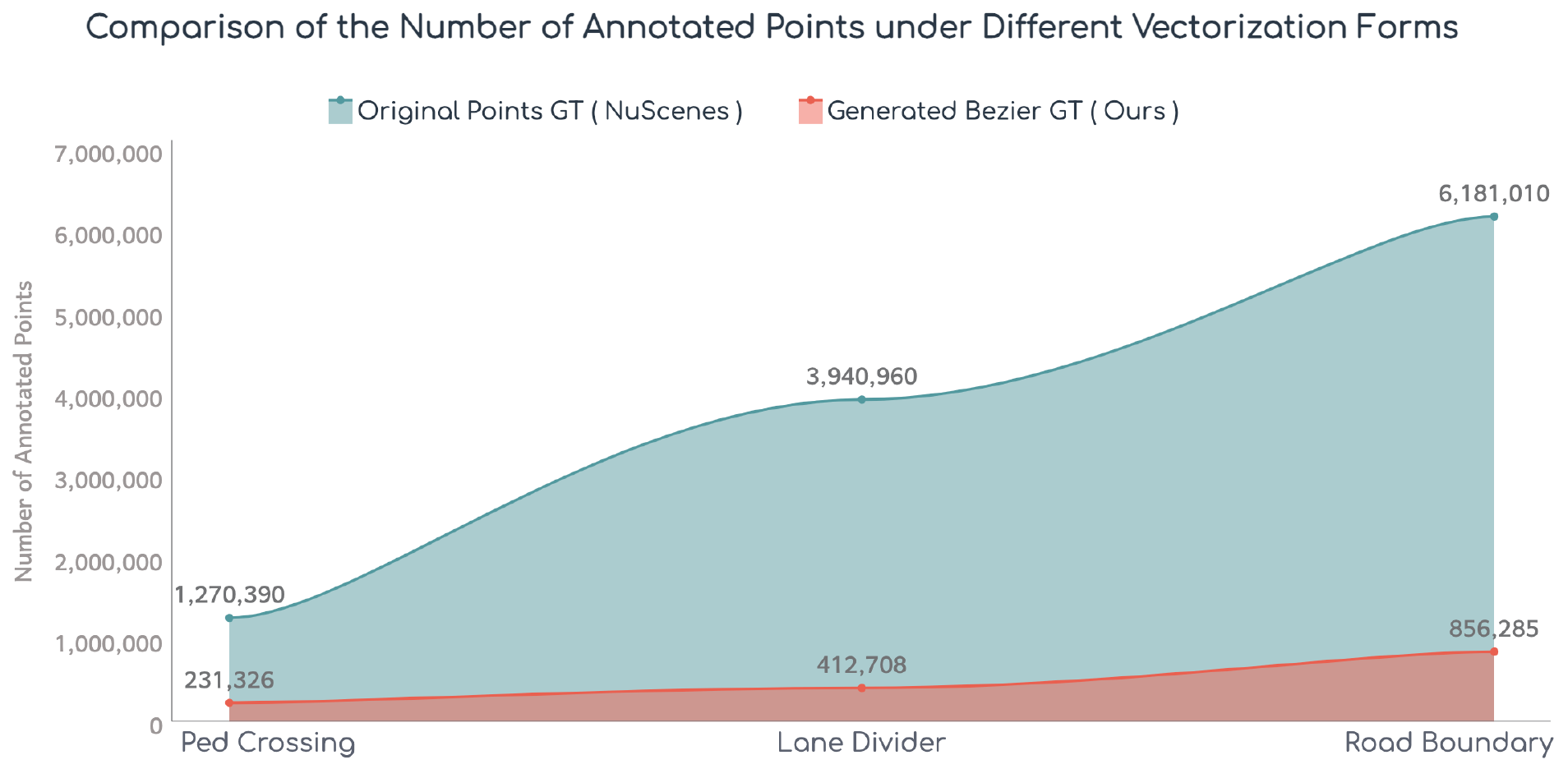}
	\end{center}
	\vspace{-0.57cm}
	\caption{The compactness of different annotated forms.}
	\vspace{-0.44cm}
	\label{fig:sat_pointnum}
\end{figure}

As for data labeling and downstream applications, the expression form of map elements is crucial, which affects the efficiency of data storage and transmission. 
Fig.\ref{fig:sat_pointnum} compares the \nuscene original point-GT and Generated \Bezier curve-GT from the perspective of the number of required annotated points.
We notice that the latter form is more than \bm{$81\%$} more effective than the former at least, indicating that \Bezier curve is a more compact expression pathway.

\subsection{How many pieces are reasonable?}
\begin{figure}[htb]
	\begin{center}
		\includegraphics[width=0.95\linewidth]{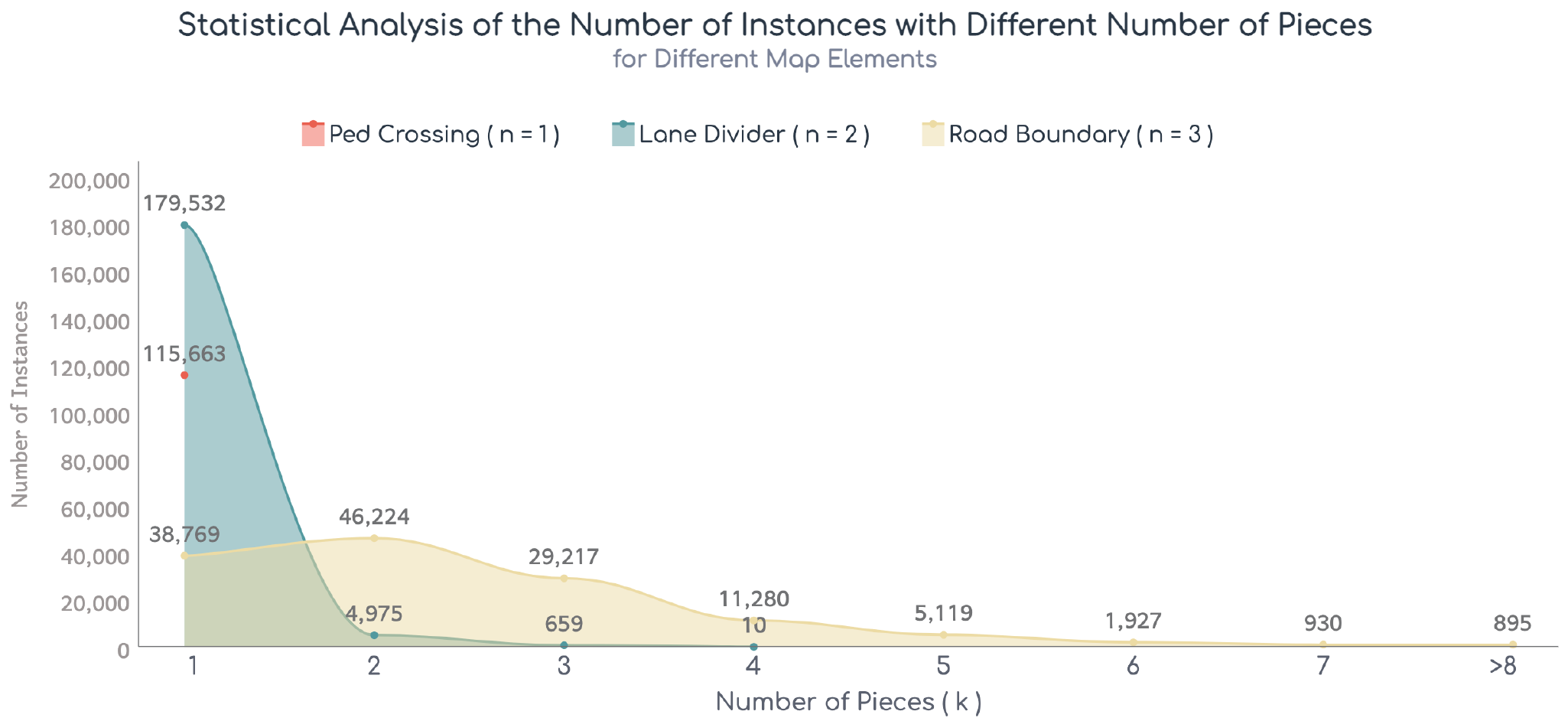}
	\end{center}
	\vspace{-0.57cm}
	\caption{
		The statistics of instance number under different pieces.
	}
	\vspace{-0.44cm}
	\label{fig:sat_piecenum}
\end{figure}

when using piecewise \Bezier curve to express \hdmap, the number of pieces for different object is various with a given degree.
Fig.\ref{fig:sat_piecenum} counts the number of segments required for different map instances in \nuscene under the prerequisite of being fully represented.
In the main paper, we utilize the deployment of \bm{$\langle 3, 2 \rangle, \langle 1, 1 \rangle, \langle 7, 3 \rangle$} for \LD, \PC and \RB respectively, which is a reasonable setting according to the Fig.\ref{fig:sat_piecenum}.

\subsection{How many degree are reasonable?}
\begin{figure}[htb]
	\begin{center}
		\includegraphics[width=0.95\linewidth]{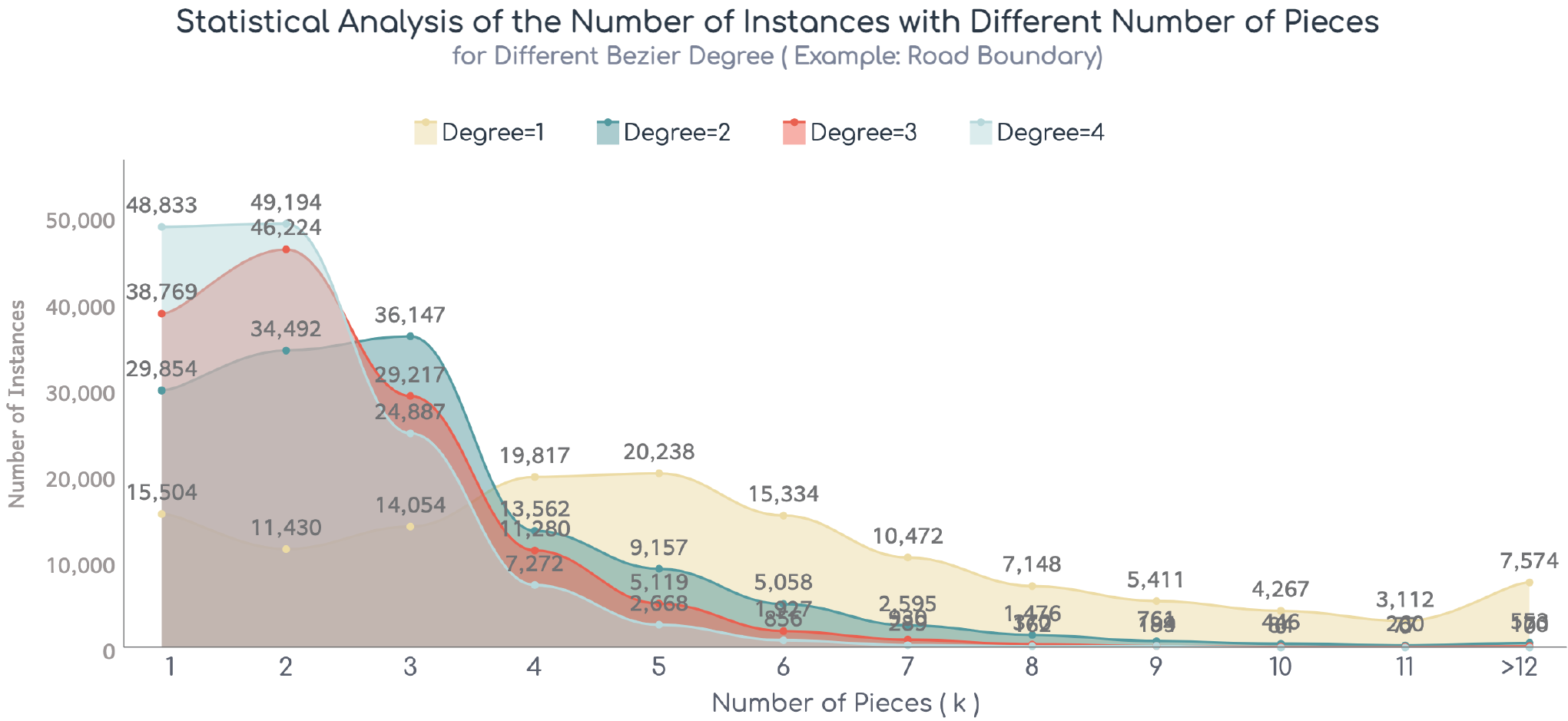}
	\end{center}
	\vspace{-0.57cm}
	\caption{
		The statistics of \textit{ins.-}number under different degree.
	}
	\vspace{-0.44cm}
	\label{fig:sat_piecenum2}
\end{figure}

Taking \RB as an example, Fig.\ref{fig:sat_piecenum2} shows the piece-number \& instance-number distribution at different degree. Note the higher the degree, the fewer the number of pieces required and the curve is more difficult to model.

\section{Qualitative Visualization}
\begin{figure*}[htb]
	\begin{center}
		\includegraphics[width=0.78\linewidth]{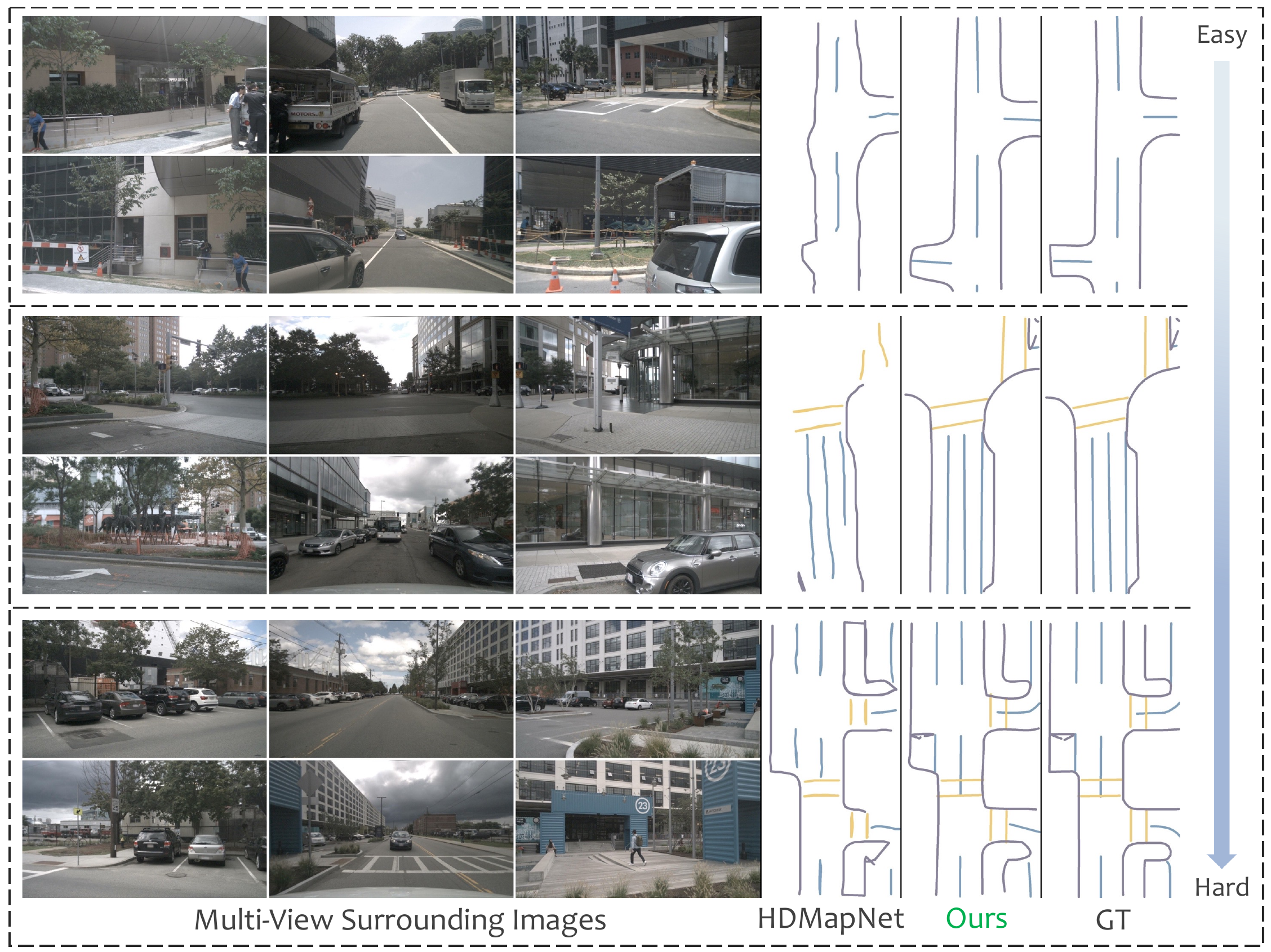}
	\end{center}
	\vspace{-0.6cm}
	\caption{
		The visualization results under the lighting condition of \textit{\textbf{daytime}} (easy $\rightarrow$ hard).
	}
	\vspace{-0.1cm}
	\label{fig:vis_day}
\end{figure*}

\begin{figure*}[htb]
	\begin{center}
		\includegraphics[width=0.78\linewidth]{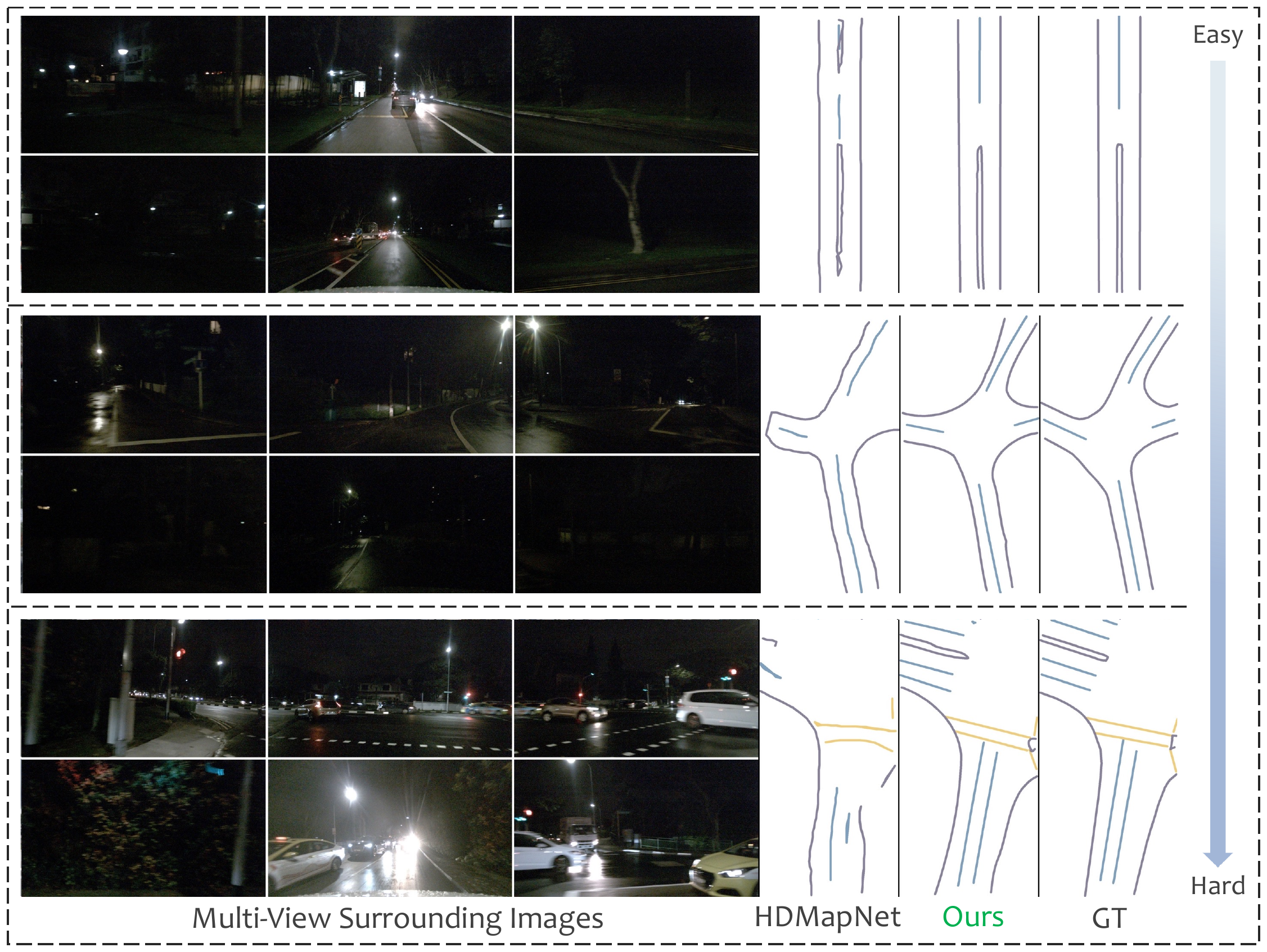}
	\end{center}
	\vspace{-0.6cm}
	\caption{
		The visualization results under the lighting condition of \textit{\textbf{nighttime}} (easy $\rightarrow$ hard).
	}
	\vspace{-0.1cm}
	\label{fig:vis_night}
\end{figure*}

\begin{figure*}[htb]
	\begin{center}
		\includegraphics[width=0.78\linewidth]{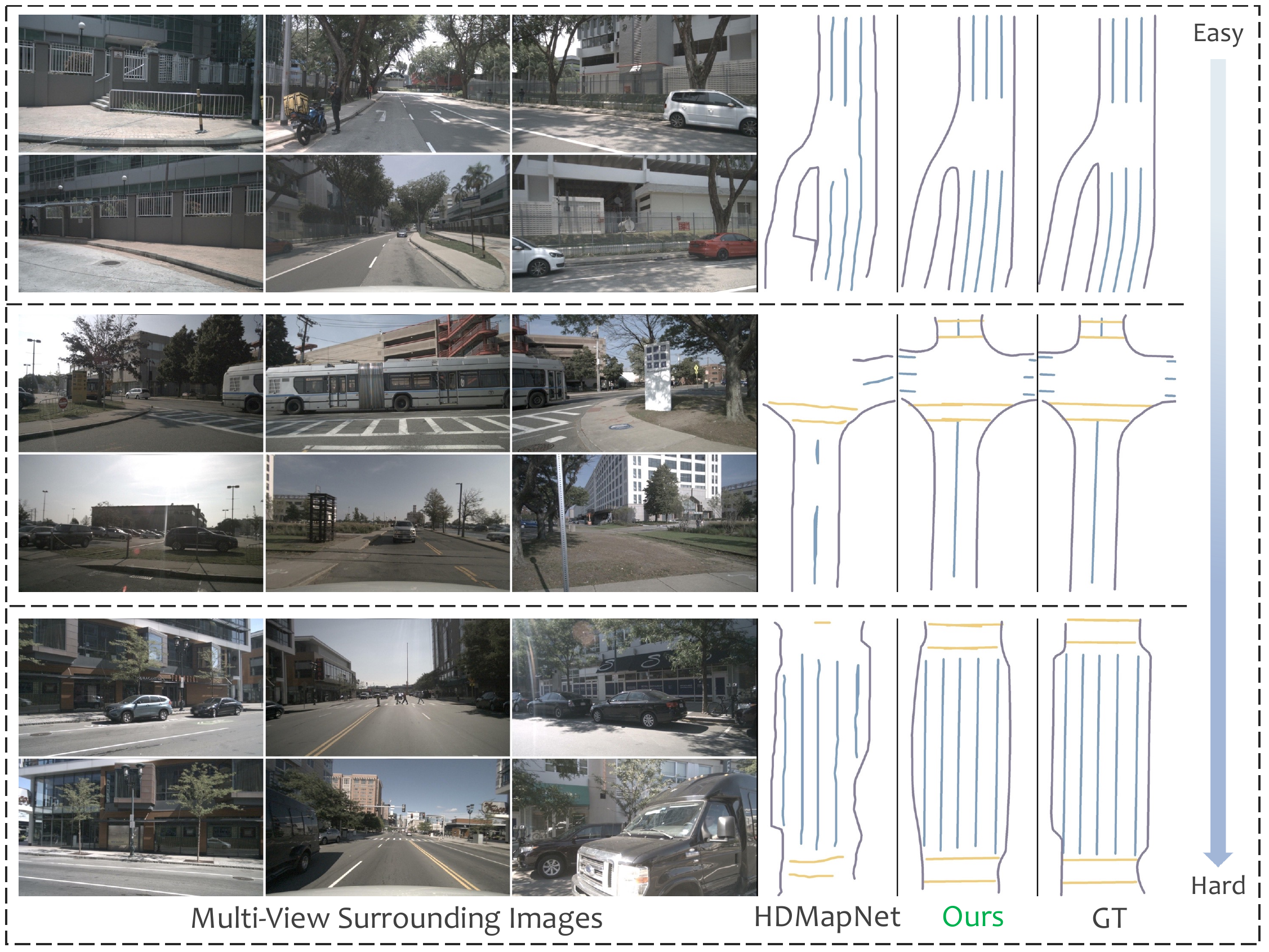}
	\end{center}
	\vspace{-0.6cm}
	\caption{
		The visualization results under the weather condition of \textit{\textbf{sunny}} (easy $\rightarrow$ hard).
	}
	\vspace{-0.1cm}
	\label{fig:vis_sunny}
\end{figure*}

\begin{figure*}[htb]
	\begin{center}
		\includegraphics[width=0.78\linewidth]{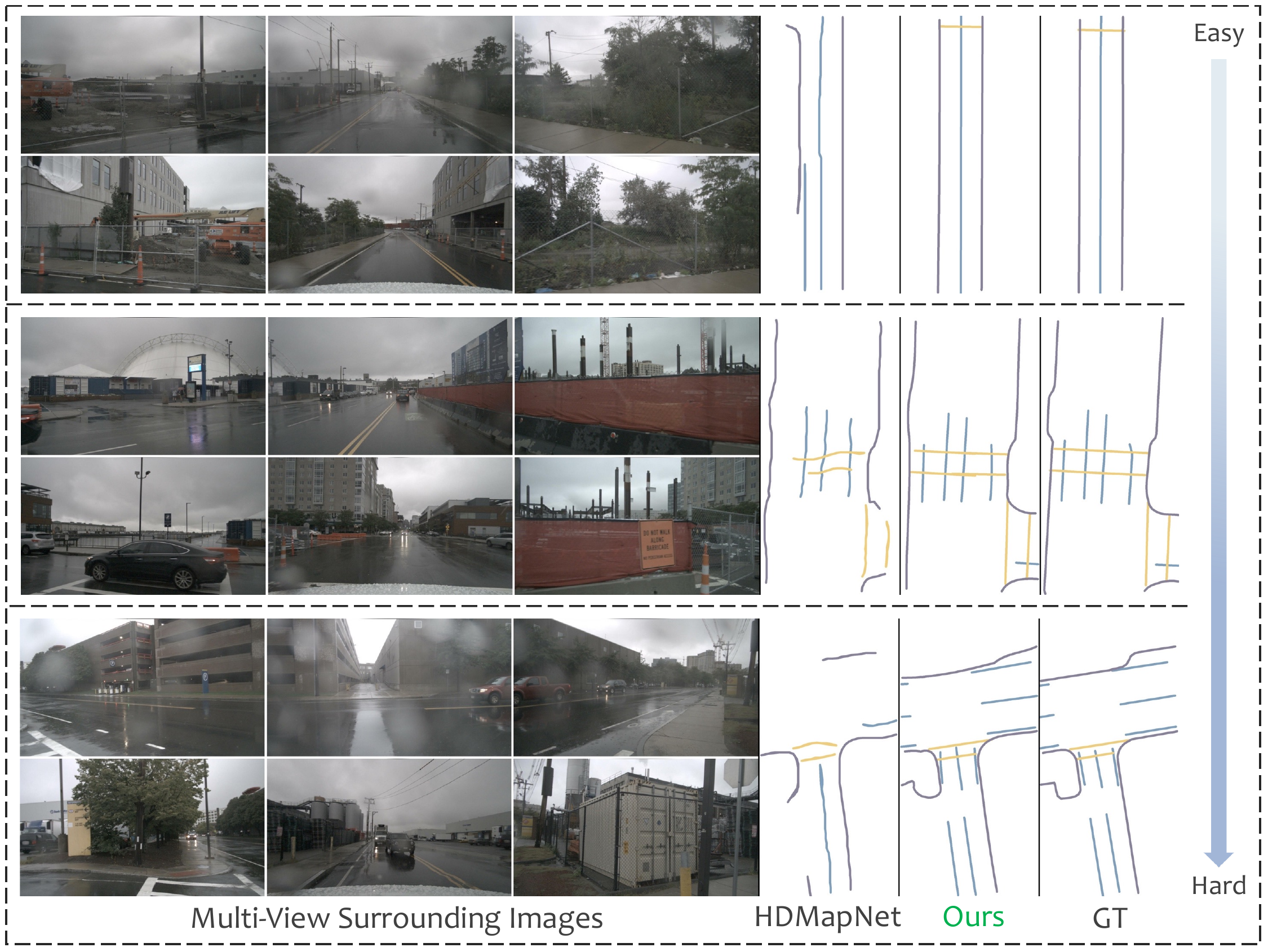}
	\end{center}
	\vspace{-0.6cm}
	\caption{
		The visualization results under the weather condition of \textit{\textbf{rainy}} (easy $\rightarrow$ hard).
	}
	\vspace{-0.1cm}
	\label{fig:vis_rainy}
\end{figure*}

\begin{figure*}[htb]
	\begin{center}
		\includegraphics[width=0.98\linewidth]{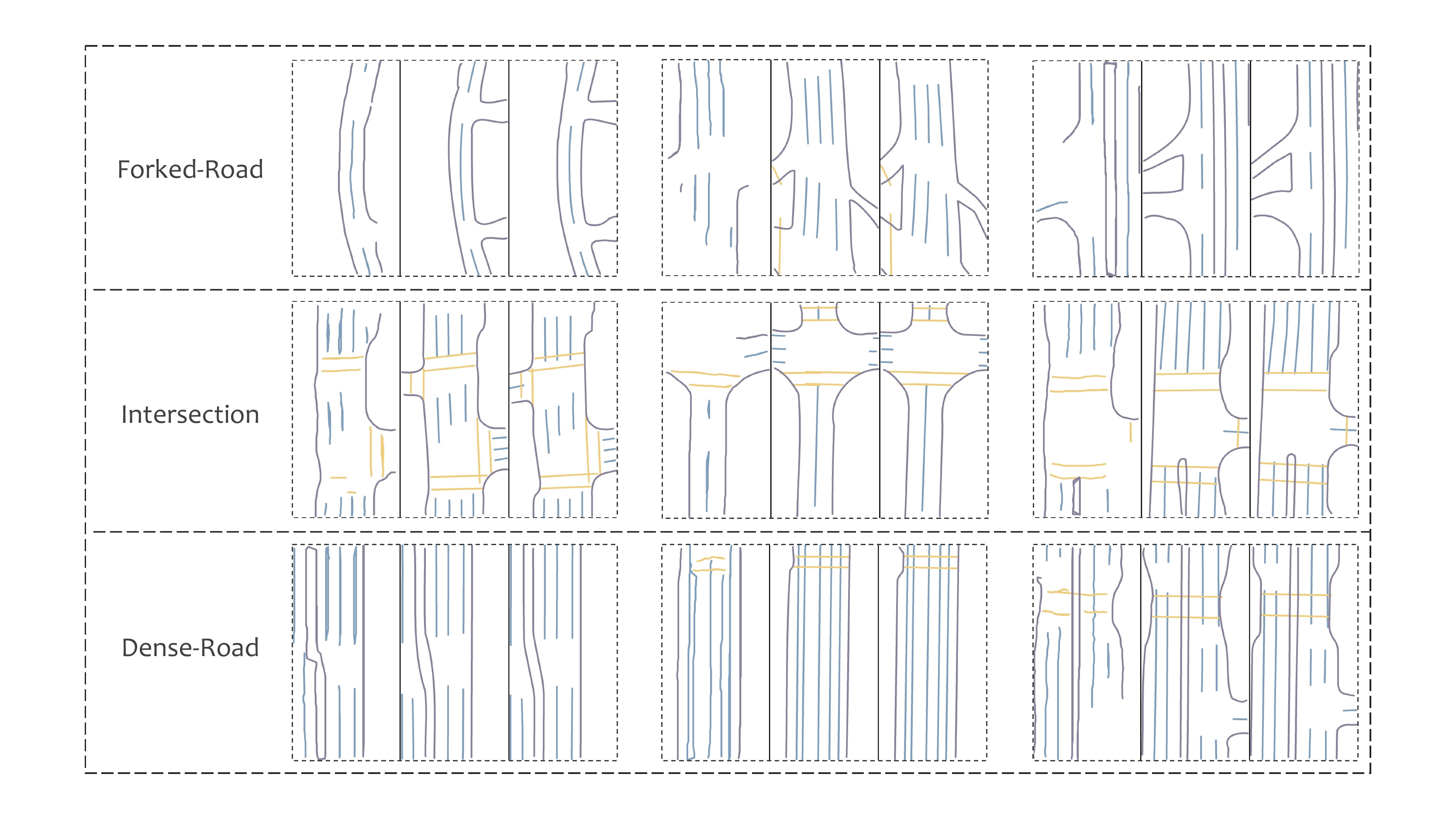}
	\end{center}
	\vspace{-0.6cm}
	\caption{
		More visualization results under difficult road scenarios (Predictions from \textit{HDMapNet}, \modelbf and Ground-Truth).
	}
	\vspace{-0.1cm}
	\label{fig:vis_other}
\end{figure*}

\begin{figure*}[htb]
	\begin{center}
		\includegraphics[width=0.78\linewidth]{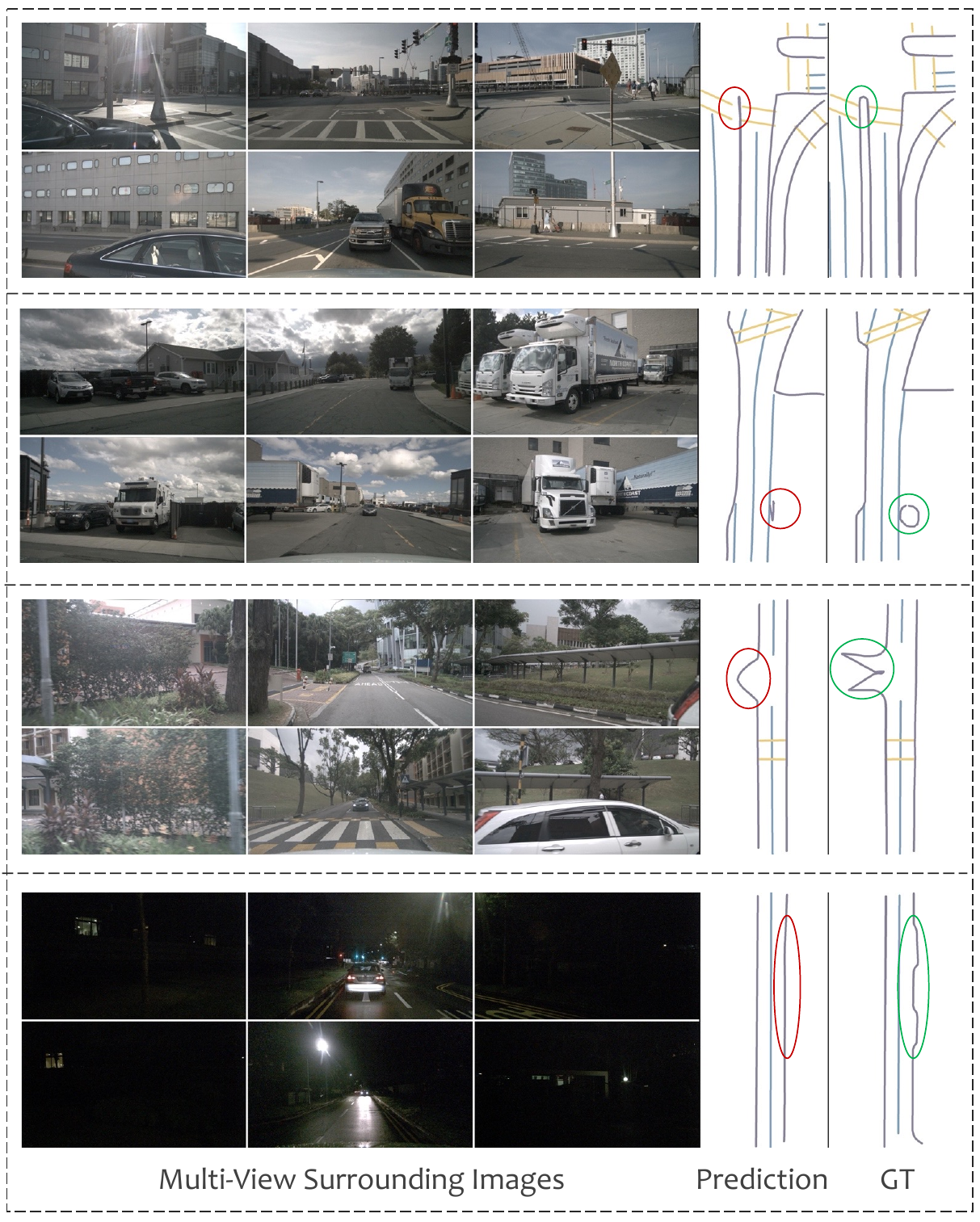}
	\end{center}
	\vspace{-0.6cm}
	\caption{
		Some badcases in our current model (for future improvement).
	}
	\vspace{-0.1cm}
	\label{fig:vis_badcase}
\end{figure*}

\begin{itemize}
	\setlength\itemsep{-0.15em}
	\item Fig.\ref{fig:vis_day}$\sim$\ref{fig:vis_night}: results under different lighting conditions.
	\item Fig.\ref{fig:vis_sunny}$\sim$\ref{fig:vis_rainy}: results under different weather conditions.
	\item Fig.\ref{fig:vis_other}: more results under difficult road scenarios.
	\item Fig.\ref{fig:vis_badcase}: some badcases for future improvement.
\end{itemize}

\subsection{The reason for rounded corner arising} 
\begin{figure}[htb]
	\begin{center}
		\includegraphics[width=0.95\linewidth]{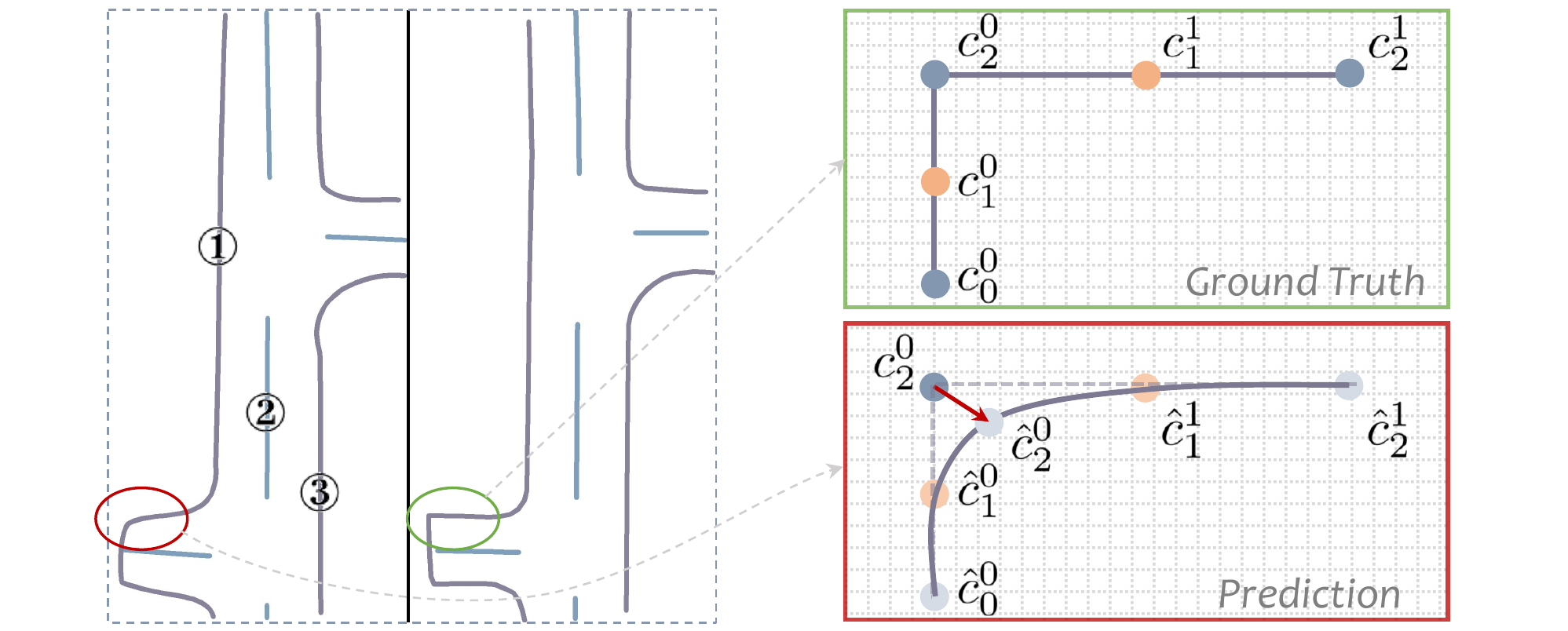}
	\end{center}
	\vspace{-0.57cm}
	\caption{The illustration for rounded corner interpretation.}
	\vspace{-0.44cm}
	\label{fig:corner_issue}
\end{figure}

The rounded corner issue is mainly caused by the slightly inaccurate prediction of some \textit{\textbf{key}} control points.
For example, a right-angle case in Fig. \ref{fig:corner_issue} can be simply formulated by  \bm{$\langle 2, 2 \rangle$} with $5$ control points in GT. Assuming that the control point prediction at the turn position has only a small offset ($c^0_2 \red{\rightarrow} \hat{c}^0_2$), while the other locations are completely accurate, the final restored \Bezier curve will naturally produce rounded corner.
Further efforts on some \textit{\textbf{key}} control points are important future jobs.

\vspace{-0.3cm}

\begin{figure*}[htb]
	\begin{center}
		\includegraphics[width=0.78\linewidth]{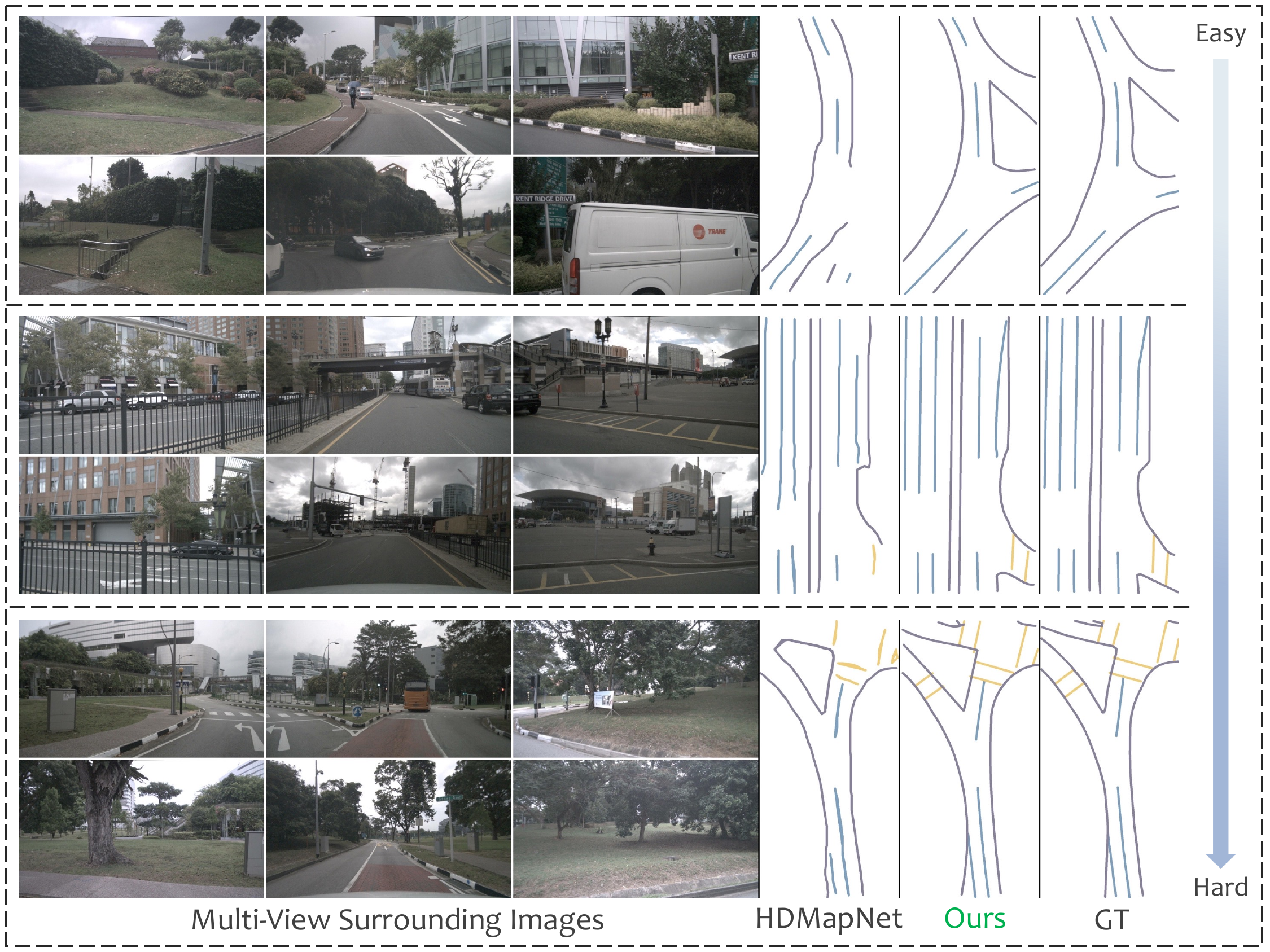}
	\end{center}
	\vspace{-0.6cm}
	\caption{
		The visualization results under the weather condition of \textit{\textbf{cloudy}} (easy $\rightarrow$ hard).
	}
	\vspace{-0.1cm}
	\label{fig:vis_cloudy}
\end{figure*} 

\clearpage
\clearpage

{\small
	\bibliographystyle{ieee_fullname}
	\bibliography{egbib}
}

\end{document}